\documentclass[lettersize,journal]{IEEEtran}
\usepackage{amsmath,amsfonts}
\usepackage{algorithmic}
\usepackage{algorithm}
\usepackage{array}

\usepackage{textcomp}
\usepackage{stfloats}
\usepackage{url}
\usepackage{verbatim}
\usepackage{graphicx}
\usepackage{cite}
\usepackage{multirow,subfigure}
\usepackage{color}
\usepackage{booktabs}
\usepackage{arydshln}
\usepackage{hyperref,MnSymbol}
\begin{document}
	
	\title{AGIQA-3K: An Open Database for AI-Generated Image Quality Assessment}
	
	\author{Chunyi Li, Zicheng Zhang, Haoning Wu, Wei Sun,~\IEEEmembership{Member,~IEEE}, Xiongkuo Min,~\IEEEmembership{Member,~IEEE},\\Xiaohong Liu,~\IEEEmembership{Member,~IEEE}, Guangtao Zhai,~\IEEEmembership{Senior Member,~IEEE}, Weisi Lin,~\IEEEmembership{Fellow,~IEEE}
		\thanks{This work was supported by the  Shanghai Pujiang Program Grant 22PJ1406800. Corresponding author: Xiaohong Liu, Guangtao Zhai.}
		\thanks{Chunyi Li, Zicheng Zhang, Wei Sun, Xiongkuo Min, and Guangtao Zhai are with the Institute of Image Communication and Network Engineering, Shanghai Jiao Tong University, Shanghai 200240, China (email: {lcysyzxdxc,zzc1998,sunguwei,minxiongkuo,zhaiguangtao}@sjtu.edu.cn)}
		\thanks{Xiaohong Liu is with the John Hopcroft Center, Shanghai Jiao Tong University, Shanghai 200240, China (email: xiaohongliu@sjtu.edu.cn)}
		\thanks{Haoning Wu and Weisi Lin are with the S-Lab, Nanyang Technological University, Singapore 639798, Singapore (email: {haoning001,wslin}@ntu.edu.sg)}
		\thanks{Manuscript received June 7, 2023.}
	}
	
	\markboth{Journal of \LaTeX\ Class Files,~Vol.~1, No.~1, Jun~2023}%
	{Shell \MakeLowercase{\textit{et al.}}: A Sample Article Using IEEEtran.cls for IEEE Journals}
	
	
	\maketitle
	
	\begin{abstract}
		With the rapid advancements of the text-to-image generative model, AI-generated images (AGIs) have been widely applied to entertainment, education, social media, etc. However, considering the large quality variance among different AGIs, there is an urgent need for quality models that are consistent with human subjective ratings. To address this issue, we extensively consider various popular AGI models, generated AGI through different prompts and model parameters, and collected subjective scores at the perceptual quality and text-to-image alignment, thus building the most comprehensive AGI subjective quality database AGIQA-3K so far. Furthermore, we conduct a benchmark experiment on this database to evaluate the consistency between the current Image Quality Assessment (IQA) model and human perception, while proposing StairReward that significantly improves the assessment performance of subjective text-to-image alignment. We believe that the fine-grained subjective scores in AGIQA-3K will inspire subsequent AGI quality models to fit human subjective perception mechanisms at both perception and alignment levels and to optimize the generation result of future AGI models. The database is released on \url{https://github.com/lcysyzxdxc/AGIQA-3k-Database}.
	\end{abstract}
	
	
	\section{Introduction}
	
	\underline{AI} \underline{G}enerated \underline{C}ontent (\textbf{AIGC}) refers to all types of content generated by artificial intelligence technology. As vision is the most important way for humans to perceive external information, AI-Generated Images (AGI), especially Text-to-Image (T2I) generation, has become one of the most representative forms of AIGC \cite{review:texttoimage}. With the rapid technological advancement of visual computing and networking, a huge variety of AGI models have emerged which include the following 3 types \cite{review:3model}. Generative Adversarial Networks \cite{review:GAN-wufeng} (GAN)-based models, such as Text-conditional GAN ~\cite{gen:TextGAN,gen:StackGAN,gen:AttnGAN} series, are the earliest end-to-end AGI model from character level to the pixel level. Since then, AGI has differentiated into two technical routes, namely auto regressive-based models ~\cite{gen:Cogview,gen:DALLE,gen:Parti} as CogView, and diffusion-based models like Stable-Diffusion ~\cite{gen:GLIDE,gen:SD,gen:XL} (SD). According to the statistics ~\cite{review:texttoimage,review:3model}, there have been at least 20 representative T2I AGI models up to 2023.
	
	With such a large number of models, the quality of AGI also varies widely \cite{review:3model}. Firstly, for different models, the GAN-based models generate AGI with the worst quality, the quality of the auto-regression-based models is comparatively improved, and the diffusion-based models generate the best results overall. Furthermore, the quality of AGIs generated by the same model can still vary greatly. For example, a large amount of training data, sufficient epoch iterations, and well-designed prompts have a huge impact on the generated result. Considering the great variance of T2I AGI content, how to fairly evaluate their quality becomes a pivotal question.
	Since the AGIs' receiving end is the Human Visual System (HVS), subjective assessment is the most direct and reliable way to quantify their quality. A fine-grained comprehensive subjective quality experiment not only helps to understand the perception mechanism for AGIs but also lay the foundation for assessing, comparing, and optimizing AGI models.
	
	\begin{figure}[tb]
		\centering
		\includegraphics[width = 0.8\linewidth]{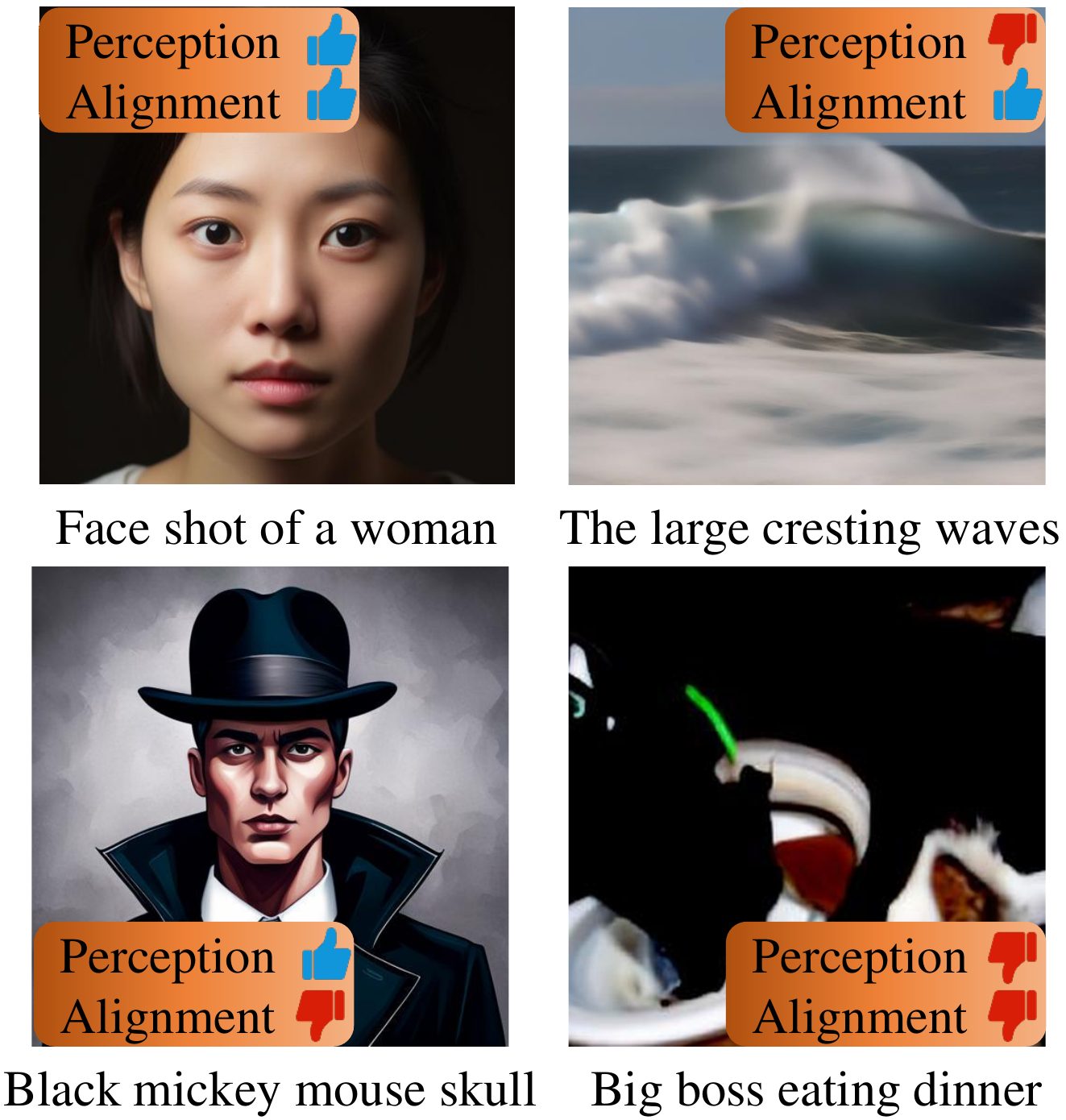}
		\caption{Illustration of the quality dimensions of AGIs: Perceptual quality and Alignment, attached with example with different subjective qualities. }
		\label{fig:spotlight}
	\end{figure}
	
	However, large-scale experiments on subjective quality assessment face two main practical challenges. Firstly, with the great diversity of T2I AGI models and the differences of AGIs themselves, it is difficult to provide fine-grained subjective scores for numerous generation models under different inputs. Therefore, the selection of generative models, parameter components, and input prompts needs to represent as wide a range of AGI as possible under a limited data scale. Secondly, considering the different properties of AGIs and NSIs, it is necessary to establish a solid subjective evaluation standard. There are several ~\cite{database:AGIQA1K, database/align:ImageReward} but no unified standards on the dimensions of AGIs' quality, and the specific information included in each dimension. Therefore, we walk through the previous AGI model, including the model itself, parameters, and prompts, set up a comprehensive set of subjective testing methods, and established the most comprehensive fine-grained, multi-dimensional AGIs quality database so far, namely AGIQA-3K as shown in Fig. \ref{fig:spotlight}. On this database, we can further explore the generation or quality model of AGIs, so as to optimize the human perception experience of AGIs. The main contributions of our work include:
	
	\begin{itemize}
		\item A large-scale AGI database consists of 2,982 AGIs generated from 6 different models. This is the first database that covers AGIs from GAN/auto regression/diffusion-based model altogether. Meanwhile, the input prompt and internal parameters in AGI models have been carefully designed and adjusted.
		\item A fine-grained subjective experiment carried out in a standardized laboratory environment. We collected the Mean Opinion Score (MOS), to annotate images in detail from both perception and T2I alignment dimensions. Therefore, we compared the result of different AGI models in different dimensions.
		\item A benchmark experiment was conducted to evaluate the performance of current perceptual quality and T2I alignment assessment metrics. Moreover, we proposed an alignment metric called StairReward to improve the existing alignment assessment result from different AGI models.
	\end{itemize}
	
	\begin{table*}[tbph]
		\centering
		\caption{Comparison of Text-to-Image AGI quality databases.}
		\label{tab:database}
		\begin{tabular}{c|c|c|c|c|c|c}
			\toprule
			& Score              & Dimension             & Image   & Ratings & Generation                          & Public Available \\ \hline
			DiffsionDB\cite{database:DiffusionDB}  & No                 & No                    & 1,819,808 & 0       & Diffusion (1)                       & \href{https://huggingface.co/datasets/poloclub/diffusiondb}{Yes}       \\ \hline
			AGIQA-1K\cite{database:AGIQA1K}    & MOS                & Perception            & 1,080    & 23,760   & Diffusion (2)                       & \href{https://github.com/lcysyzxdxc/AGIQA-1k-Database}{Yes}        \\ \hline
			Pick-A-Pic\cite{database/align:PickAPic}  & Preference         & Overall               & 500,000  & 500,000  & Diffusion (3)                       & \href{https://huggingface.co/datasets/yuvalkirstain/pickapic_v1}{Yes}        \\ \hline
			HPS\cite{database/align:HPS}        & Preference         & Overall               & 98,807   & 98,807   & Diffusion (1)                       & \href{https://mycuhk-my.sharepoint.com/personal/1155172150_link_cuhk_edu_hk/_layouts/15/onedrive.aspx?id=%2Fpersonal%2F1155172150%5Flink%5Fcuhk%5Fedu%5Fhk%2FDocuments%2FHPS%2Fdataset%2Ezip&parent=%2Fpersonal%2F1155172150%5Flink%5Fcuhk%5Fedu%5Fhk%2FDocuments%2FHPS&ga=1}{Yes}        \\ \hline
				ImageReward\cite{database/align:ImageReward} & Seven Point Likert & Perception; Alignment & 136,892  & 410,676  & Auto Regressive; Diffusion (6)      & No        \\ \hline
				AGIQA-3K    & MOS                & Perception; Alignment & 2,982    & 125,244  & GAN; Auto Regressive; Diffusion (6) & \href{https://github.com/lcysyzxdxc/AGIQA-3k-Database}{Yes}      \\ \bottomrule
			\end{tabular}
		\end{table*}
		
		\section{Related Work}
		
		\subsection{AGI Quality Metric}
		\label{sec:quality}
		For AGIs' quality assessment, perceptual quality and T2I alignment \cite{review:3model} have always been the two major components. In the perspective of perceptual quality, Inception Score (IS) \cite{quality:is} is the earliest quality criterion by calculating the uniformity of a set of AGIs' features. Subsequently, methods such as Fréchet Inception Distance (FID) \cite{quality:fid} and Kernel Inception Distance (KID) \cite{quality:kid} appeared, which use the distance between the AGIs group and the NSIs group to represent the perceptual quality. However, the above methods are usually only suitable for evaluating the quality of a group of images (E.g. The performance of an AGI model) or style transfer \cite{quality:artness-luojiebo}, which is unsuitable for evaluating the perceptual quality of only one image. Therefore, for a single AGI's quality, the Image Quality Assessment (IQA) \cite{review:iqa} method is usually used. However, considering the complexity of AGI models and the diversity of factors affecting AGIs' quality, factors affecting the quality of AGIs \cite{database:AGIQA1K} are different from those of NSIs \cite{group1:aspect,group1:lightVQA,group1:vdpve,group1:wild}, and Screen Content Images (SCIs) \cite{group1:screen,tcsvt:screen,group1:cartoon}. Thus, the reliability of IQA measures is also limited. 
		
		When it comes to T2I alignment, several metrics represented by Contrastive Language-Image Pre-Training (CLIP) \cite{align:clip, quality/align:XIQE,other:clipvip} are widely applied. Those metrics can link the text with the image, which trades off against quality metrics to provide guidance for image generation. Unfortunately, the great difficulty of training these alignment models has resulted in most users may only load their pre-trained parameters and bee to tune on the small-scale database. Therefore, for the diverse morpheme composition of the prompts in the AGI database, the consistency between the alignment result and human subjective rating still needs to be improved.

		\subsection{AGI Quality Database}
		
		The popularity of the T2I AGI model in recent years has spawned several related databases as shown in Tab. \ref{tab:database}.
		
		DiffusionDB \cite{database:DiffusionDB} is the earliest database for AGIs, including 1.8+ millions of Text-Image pairs generated by the Stable-Diffusion model. Although it has no subjective scoring, its large number of images and prompts have laid the foundation for subsequent subjective databases.
		
		AGIQA-1K \cite{database:AGIQA1K} is the first subjective database for perceptual AGI quality assessment that conducted fine-grained scoring through MOS. Its input only contains 180 prompts, and these prompts are just simple combinations of image labels from the real world, which is difficult to represent a wide range of AGIs.
		
		Pick-A-Pic \cite{database/align:PickAPic} and HPS \cite{database/align:HPS} further expand the scale of image and prompts, which crawls the results generated by Stable-Diffusion on the Discord website or directly applies the Text-Image pairs in DiffusionDB and give a subjective score, but only using diffusion-based model leads to a limitation of representing various AGIs. Moreover, they combined the perception and alignment together with an overall score, which fails to characterize the AGIs' quality of multiple dimensions.
		
		ImageReward \cite{database/align:ImageReward} is a closed-source AGI quality database. For image generation, in addition to generating with four diffusion-based models, it also considers an auto-regression-based model. It also performed a better subjective test by scoring the AGI from 0 to 7.
		However, the absence of the GAN-based model and only using four excellent diffusion-based models lead to insufficient coverage of AGI quality; the granularity of scoring is discrete, and each picture only contains one person's scoring, so this kind of coarse-grained score cannot accurately characterize the quality of AGI.
		
		Suffering from the problems mentioned above, a fine-grained AGI quality database for both perception and alignment is needed. The above issue motivates us to build a new database for AGI perception and generation in the future, which aims to cover more AGI models in different performances/parameters and to provide more accurate quality results by further refining the scoring granularity.
		
		\begin{figure*}[thbp]
			\centering
			\includegraphics[width = \linewidth]{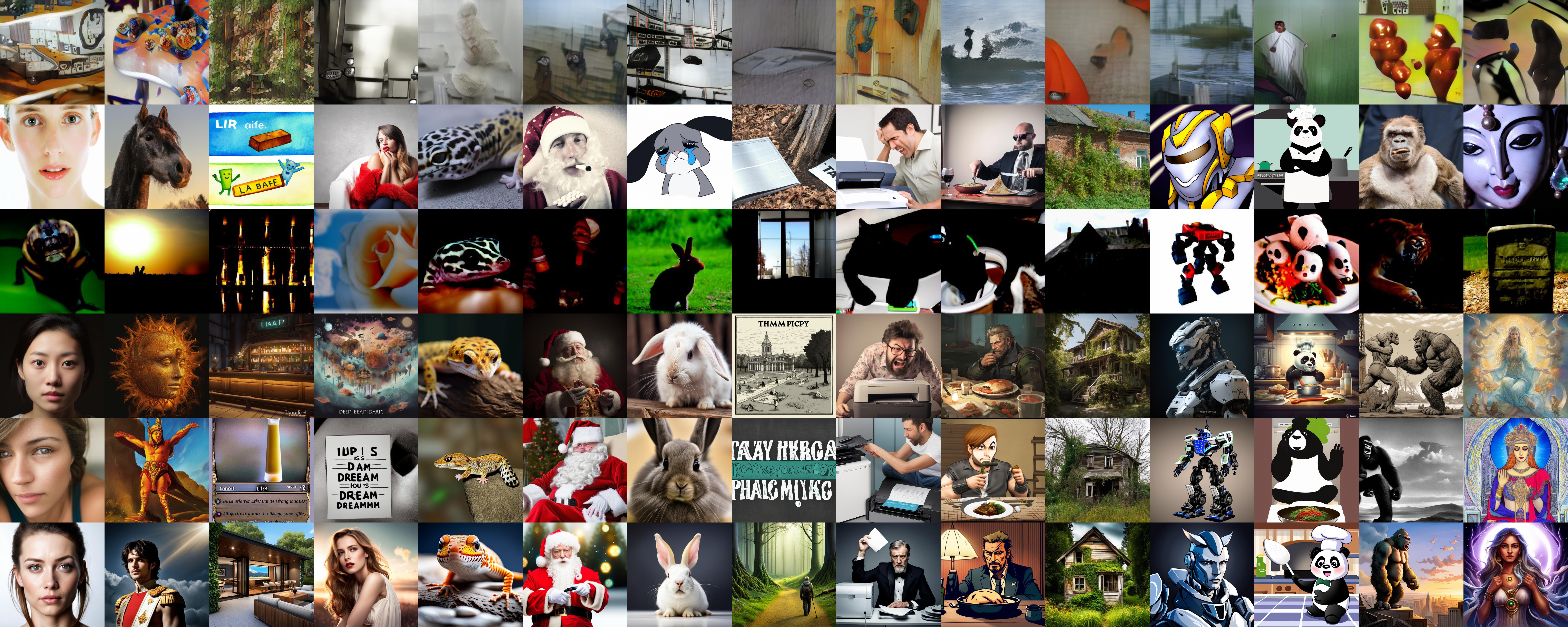}
			\caption{Sample images from the AGIQA-3K database, where the first to sixth rows show AGIs created by (\emph{AttnGAN \cite{gen:AttnGAN}, DALLE2 \cite{gen:DALLE2}, GLIDE \cite{gen:GLIDE}, Midjourney \cite{gen:MJ}, Stable Diffusion \cite{gen:SD} and Stable Diffusion XL \cite{gen:XL}}) while the column indicates the same input prompt respectively. }
			\label{fig:exhibition}
		\end{figure*}
		
		\section{Database Construction}
		
		\subsection{AGI Model Collection}
		
		\begin{figure}[thbp]
			\centering
			\subfigure[NSI distributions \cite{database:KonIQ-10k}]{\includegraphics[width = 0.40\textwidth]{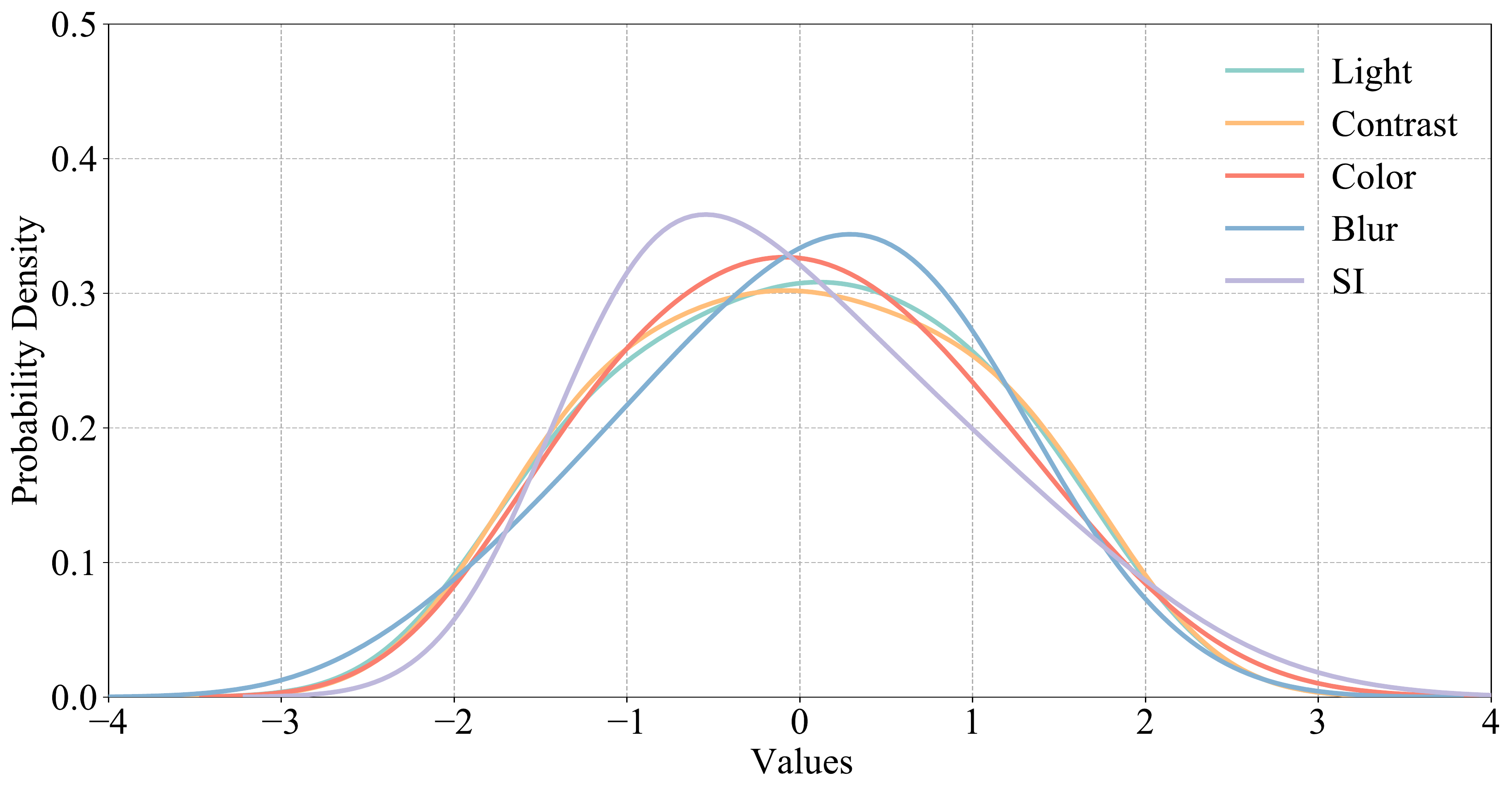}}
			\subfigure[AGIQA-1K distributions \cite{database:AGIQA1K}]{\includegraphics[width = 0.40\textwidth]{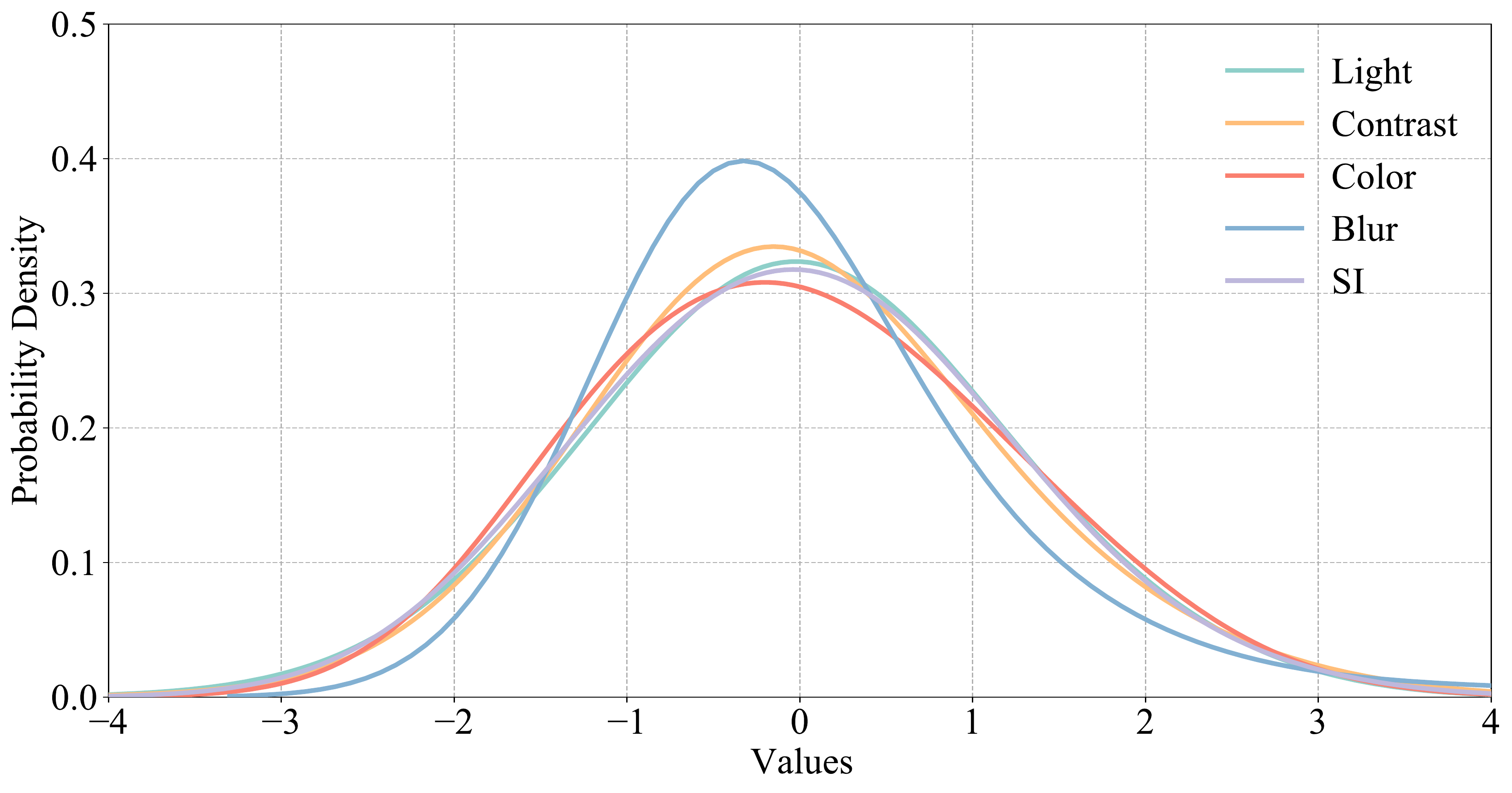}}
			\subfigure[AGIQA-3K distributions]{\includegraphics[width = 0.40\textwidth]{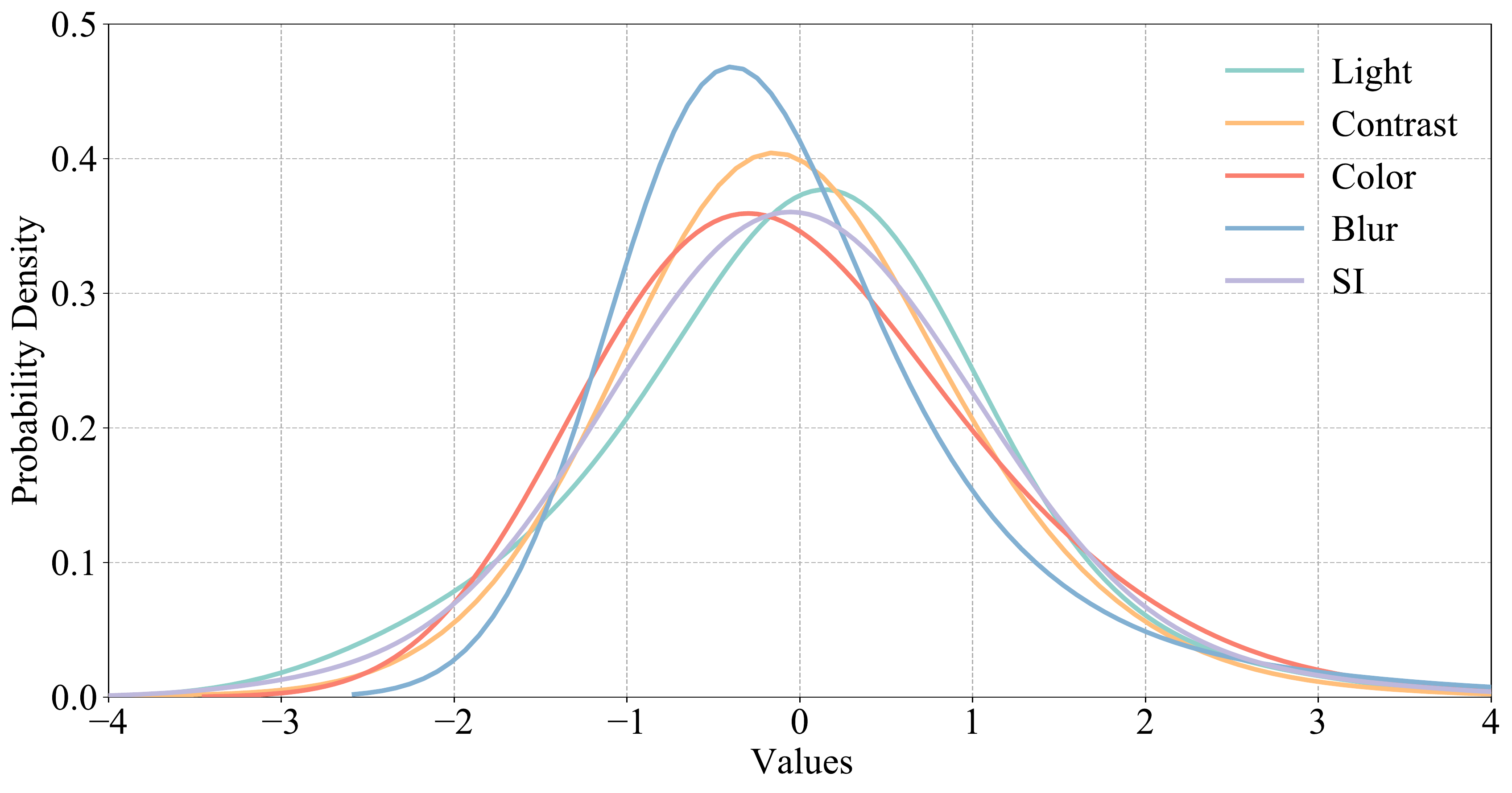}}
			\caption{The normalized probability distributions of the quality-related attributes. The distributions include NSIs in the KonIQ-10k \cite{database:KonIQ-10k}, previous AGIs in the AGIQA-1K \cite{database:AGIQA1K} and latest AGIs in the proposed AGIQA-3K database. AGIs and NSIs are similar in most visual features, but the distribution of Blur is uneven; and due to the consideration of some low-quality AGIs, the unevenness of AGIQA-3K is more significant than that of AGIQA-1K.
			}
			\label{fig:diff}
			\vspace{-6mm}
		\end{figure}
		To ensure content diversity, our AGIQA-3K database considered six representative generative models. Referring to the previous classification, considering that the overall generation effect of the diffusion-based model on the T2I AGI task is the best and the most widely used, we selected four diffusion-based models for image generation. Including the earliest GLIDE \cite{gen:GLIDE}, the most popular Stable Diffusion V-1.5 \cite{gen:SD} with its latest upgraded Beta version named Stable Diffusion XL-2.2 \cite{gen:XL}, and the highest-rated Midjourney \cite{gen:MJ} \footnote{Midjourney hasn't released their internal structure, but it's generally believed as diffusion. \cite{review:sdmjde}} by the user community. At the same time, to consider the other two types of models, we used the most popular frameworks of these two types, namely AttnGAN \cite{gen:AttnGAN} representing the GAN-based model, and DALLE2 \cite{gen:DALLE2} \footnote{DALLE-2 also applied some diffusion mechanism.} as the Auto regressive-based model. Fig. \ref{fig:exhibition} shows the output of these six AGI models. In addition, to study the relationship between the internal parameters of the model and AGI, we take the Classifier Free Guidance (CFG) scale of the AGI model as 0.5 and 2 times the default value, and explore the impact of the trade-off between perception and alignment on the generation effect; meanwhile, we set the number of iterations to half of the default value to simulate the distortion of AGI when the iterations are insufficient. These two parameter adjustments are performed on the most widely used Stable Diffusion \cite{gen:SD} and Midjourney \cite{gen:MJ}respectively. It can be seen that the AGIQA-3K database uses different models in different periods that effectively represent the wide quality range of AGI since the birth of the T2I AGI model.
		
		To assess the statistical difference between NSIs and AGIs, we propose distributions of five quality-related attributes for comparison. NSI was obtained from the KonIQ-10k IQA database \cite{database:KonIQ-10k} in the wild, while AGI was collected via the previous AGIQA-1K \cite{database:AGIQA1K} database and the proposed AGIQA-3K database. The quality-related attributes under consideration are lighting, contrast, color, blur, and spatial information (SI). The `color' indicates the colorfulness of the images and the SI stands for the content diversity of the images. A detailed description of these properties can be found in \cite{database:KoNViD-1k}. As shown in Fig. \ref{fig:diff}, there is a noticeable difference between NSIs and AGIs in the blur distribution curve because AGIs sometimes encounter insufficient iterations during the generation process. Frequent occurrence of blur causes the center of the blur distribution curve shiftting to the left. Compared with AGIQA-1K, our AGIQA-3K further adds data with insufficient model iterations, making the distortion distribution curve sharper. Except for Blur, the similarity of the distributions between four other quality-related attributes of NSIs and AGIs prove the plausibility of our AGI database. 
		
		\begin{figure}[t]
			\centering
			\includegraphics[width = 0.5\textwidth]{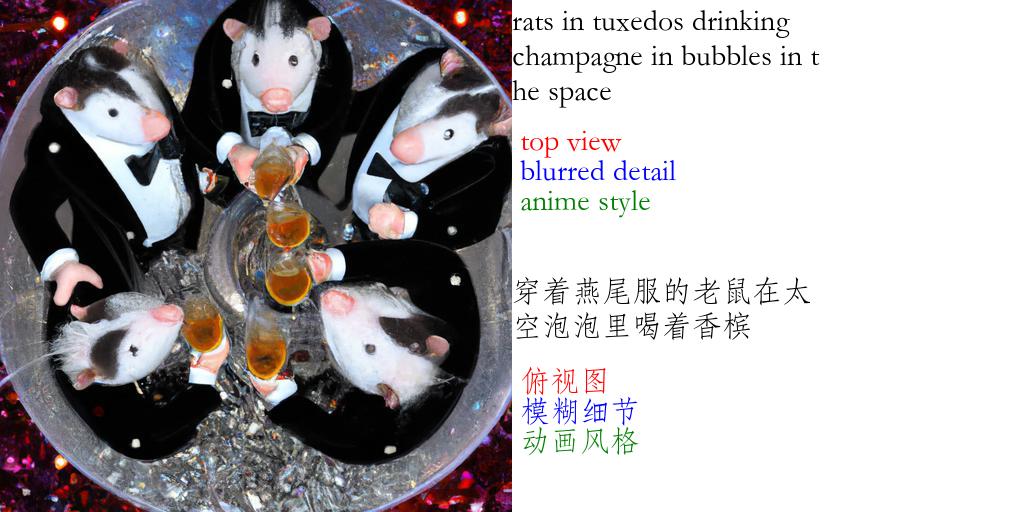}
			\caption{An example of the quality assessment interface with the AGI and corresponding prompt together. (A translation of the prompt is attached based on the viewer's nationality)}
			\label{fig:gui}
		\end{figure}
		
		
		\subsection{AGI Prompt Collection}
		
		Under the requirements of Fine-grained scoring, the AGIQA-3K database cannot perform image generation and scoring tasks on more than ten thousand prompts like the previous coarse-grained database ~\cite{database/align:PickAPic, database/align:HPS}. Therefore, how to use relatively few prompts to cover a large number of real user inputs is a key issue in the prompt collection process. Due to the insufficient prompt, directly extracting a part of the prompt from the real input will inevitably lead to one-sidedness. Facing this challenge, AGIQA-1K \cite{database:AGIQA1K} extracted high-frequency words from the Internet and created 180 Human designed prompts. However, its high-frequency words are not directly derived from the prompt input in the AGI task, and the combination of high-frequency words is also different from the common prompt input format.
		
		Therefore, our prompts in the AGIQA-3K database apply a `real' + `human designed' mechanism that uses real prompts in AGI as a framework and combines them together manually. Conform to the prompt structure of the Stable Diffusion official prompt book \footnote{https://openart.ai/promptbook}, we divide the prompt into three items, namely subject, detail, and style. The subject is the most important item that exists in all prompts. We extract 300 subjects from the prompts of DiffusionDB \cite{database:DiffusionDB} according to the proportion of the respective categories (E.g. People, Arts, Outdoor Sense, Artifacts, Animals, etc.) in ImageReward \cite{database/align:ImageReward}. Detail refers to the adjectives added after the main object of the prompt, usually no more than two. We select the ten most commonly used adjectives with reference to the real input \footnote{https://docs.google.com/spreadsheets/d/1GuAeSFtICsjQEwsRP2f--IayDxW9Dl0SCLOVov56FMc} of Midjourney users. In terms of the artistic style of the entire picture, we also selected the five most commonly used styles like the detail item. Finally, we combine the subject, 0-2 details, and 0-1 style together as shown in Fig. \ref{fig:gui}. Thus, we ensure that the prompts in AGIQA-3K cover a wide range of input content of the T2I generation task.
		
		\begin{figure}[t]
			\centering
			\subfigure[Technical issues]{\includegraphics[width = 0.23\textwidth]{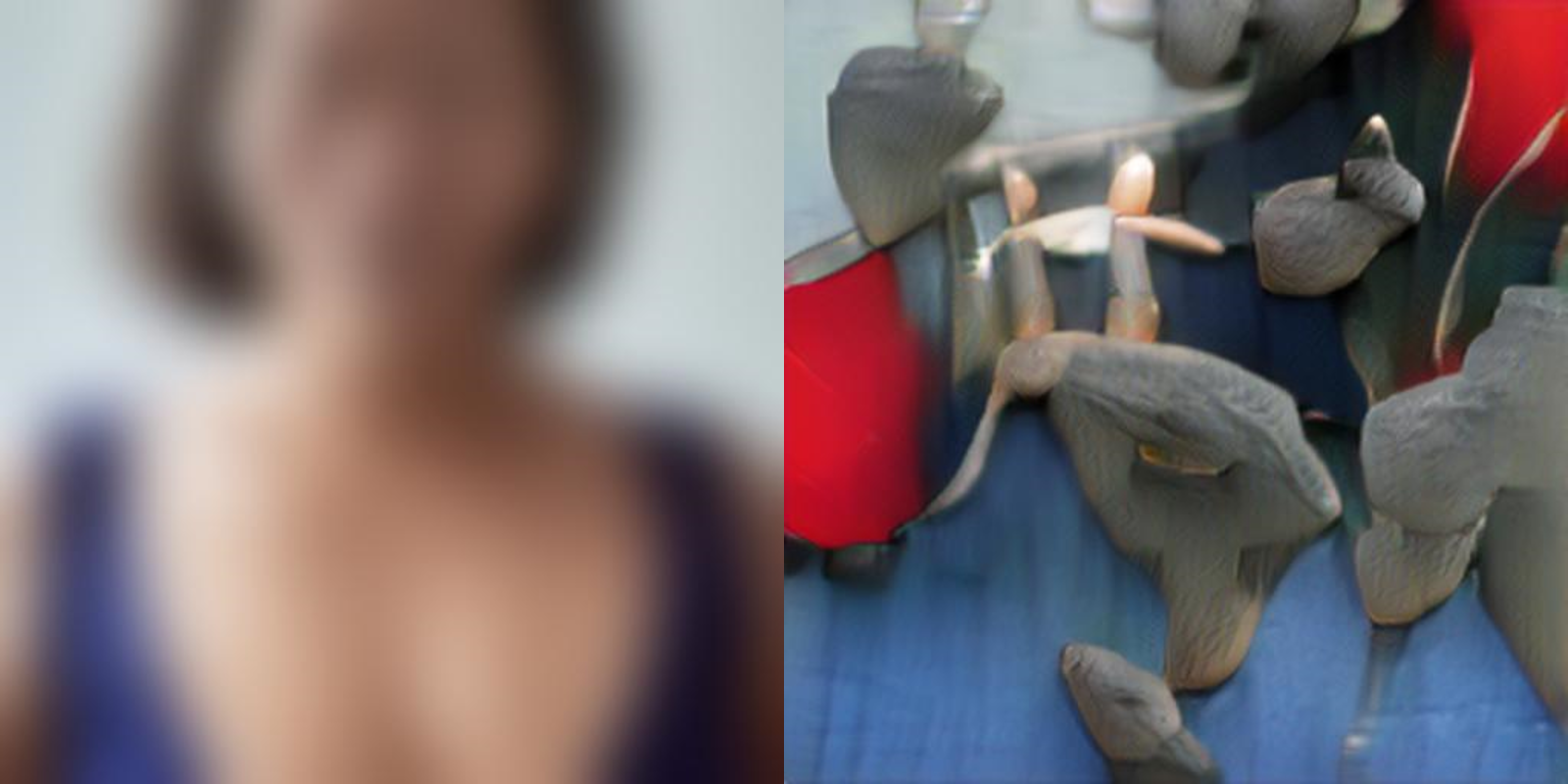}
			}
			\subfigure[AI artifacts]{\includegraphics[width = 0.23\textwidth]{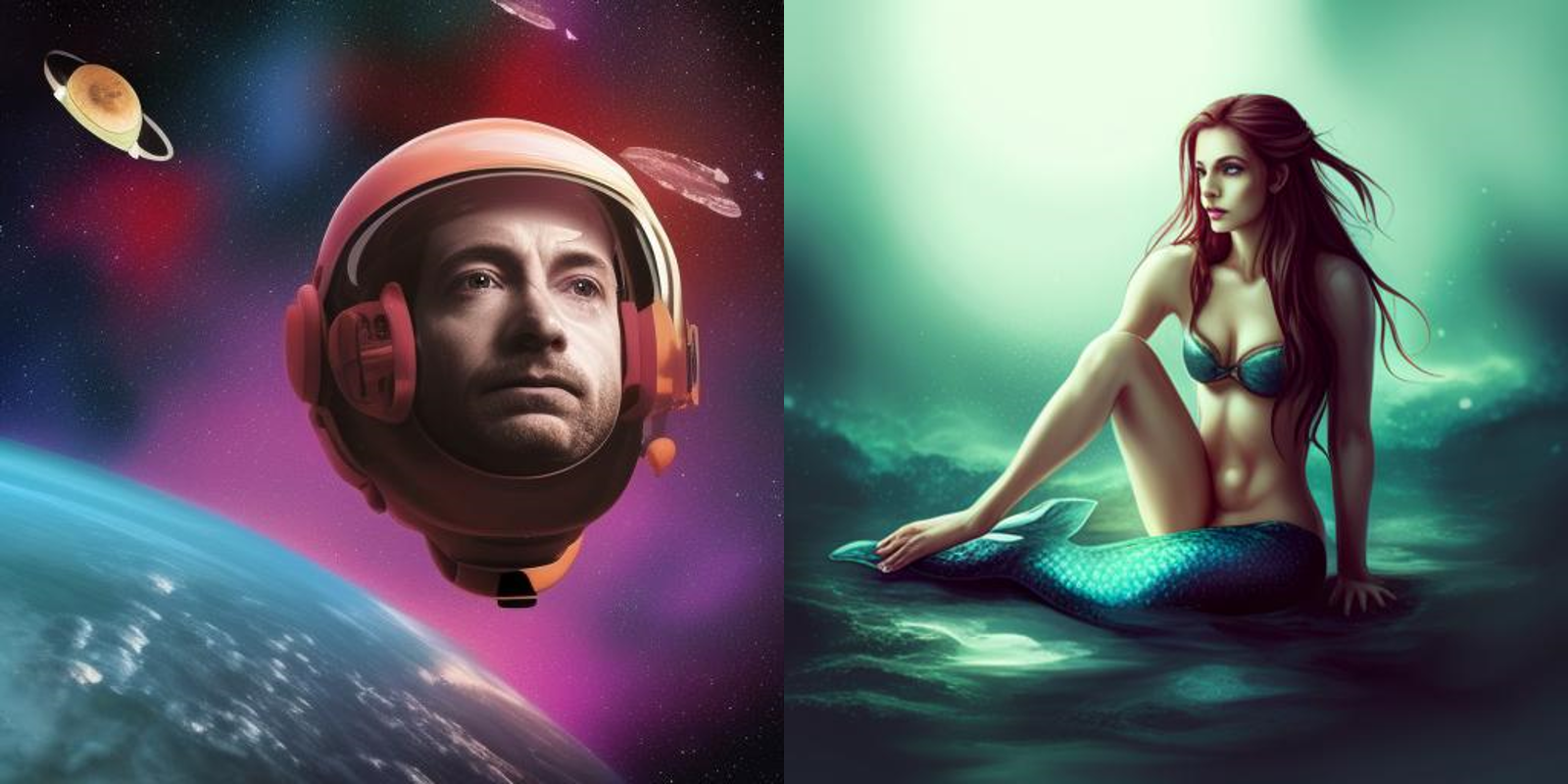}
			}
			\subfigure[Deepfake]{\includegraphics[width =  0.23\textwidth]{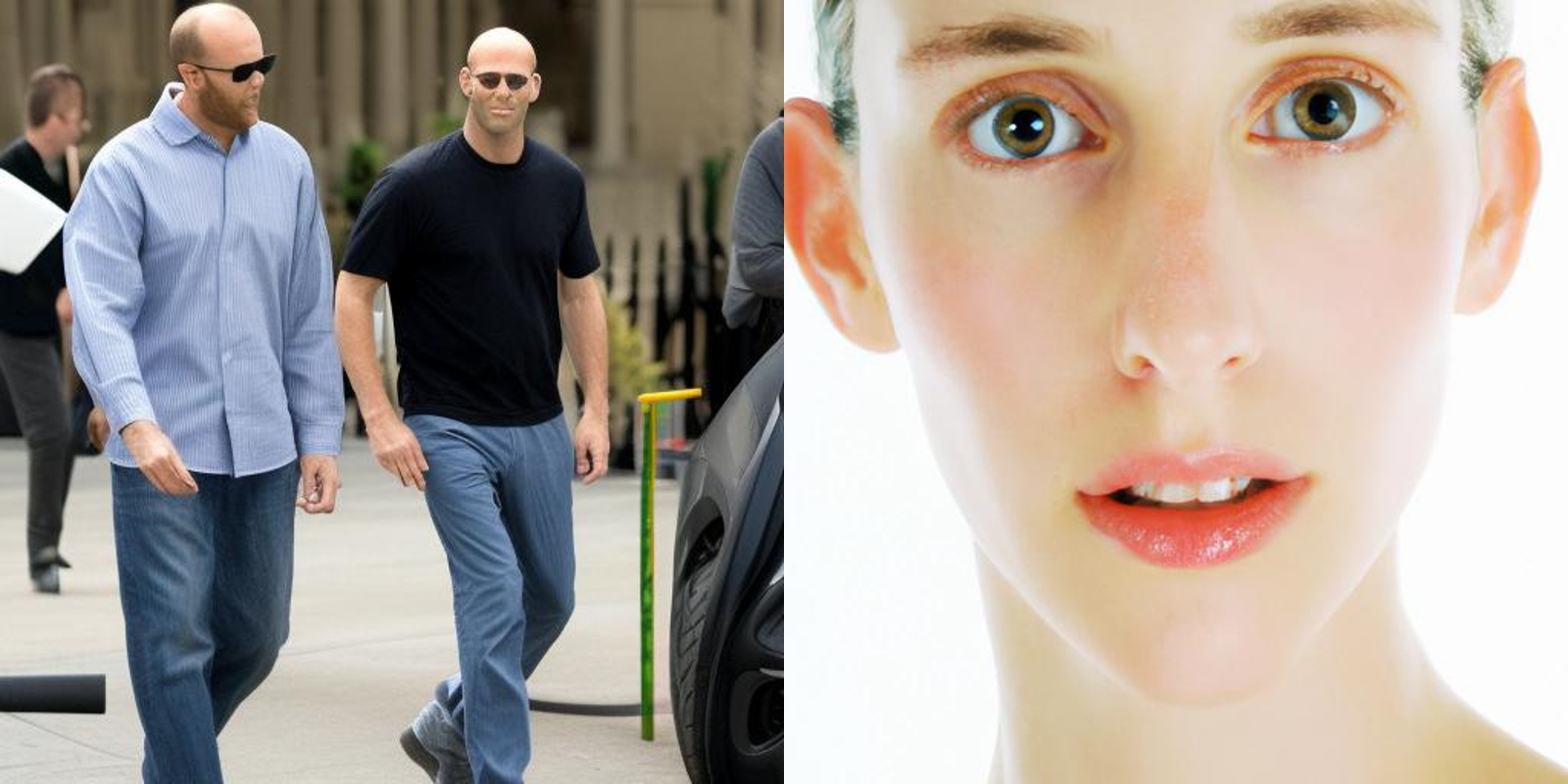}
			}
			\subfigure[Aesthetic aspects]{\includegraphics[width =  0.23\textwidth]{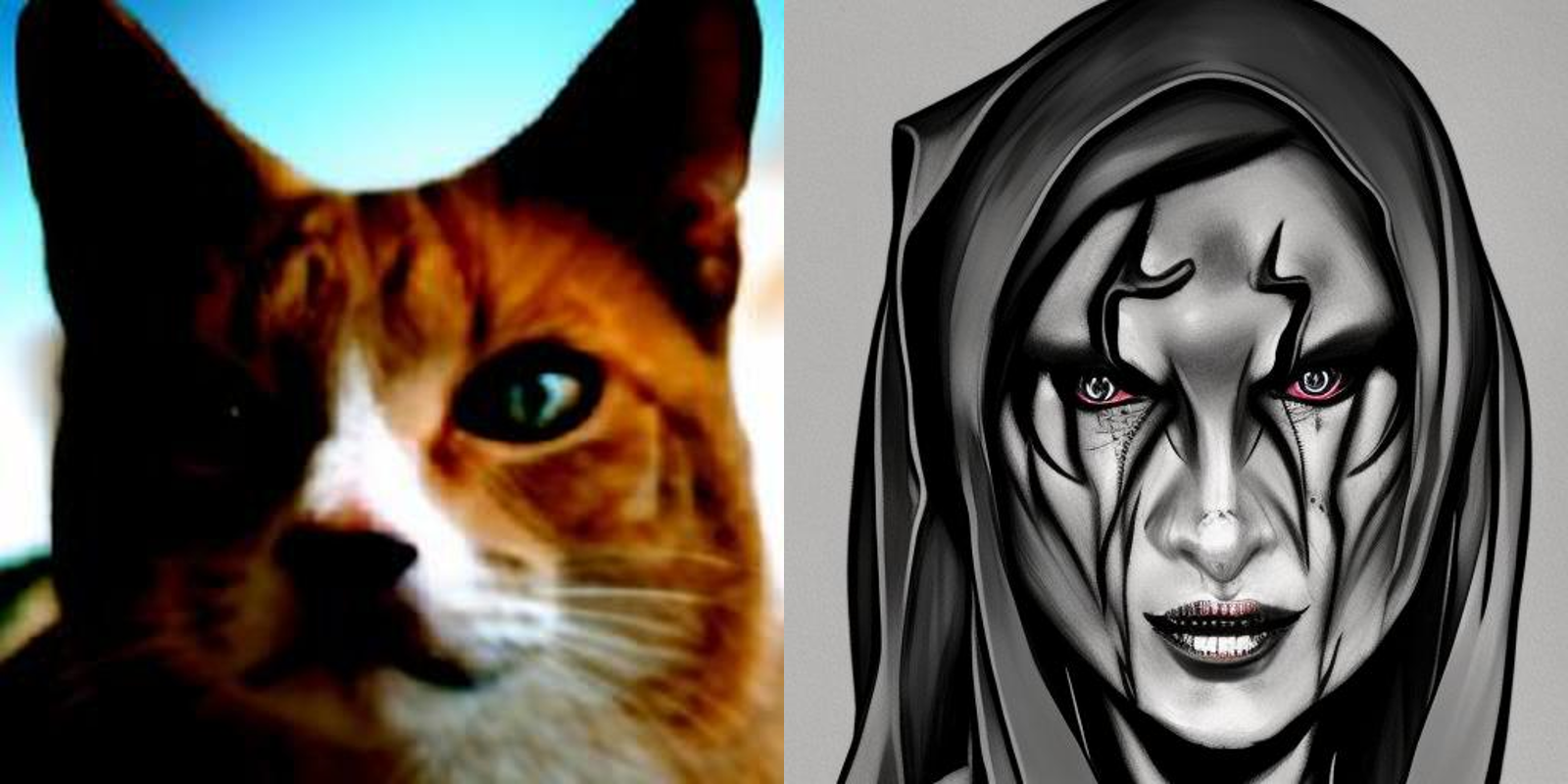}
			}
			\caption{Exhibition for some common AGI distortions, where the generation keywords are marked in the top right. (a) Technical issues are the low-level distortion like a blur and meaningless objects. (b) AI artifacts mean unexpected artifacts such as missing/excess limbs. (c) Deepfake refers to AGIs being recognized as a fake image by HVS because of unreal proportions, lighting \cite{other:deepfake}, etc. (d) Aesthetic aspects indicate the aesthetic quality including lack of detail, uncomfortable objects, etc.}
			\label{fig:distortions}
		\end{figure}
		
		\subsection{Subjective Experiment}
		To obtain the subjective quality score of AGIQA-3K, we conducted a one-month subjective experiment in the SJTU Multimedia Laboratory. Complying with the ITU-R BT.500-13 \cite{other:itu} standard, we set up the environment as a normal indoor home setting, with normal lighting levels, presented AGIs on the screen in random order on the iMac monitor with a resolution of up to 4096 $\times$ 2304 interface, and score them at Perception and Alignment level through two sliders. The perception score is an overall score of technical issues, AI artifacts, deepfake, and aesthetic aspects. Some typical distortion examples are shown in Fig. \ref{fig:distortions}. The alignment score \cite{tcsvt:T2ialignment1,tcsvt:T2ialignment2} stands for the compatibility between AGIs and prompts (including all items in the prompt, generally the subject is more critical than detail and style).
		
\begin{figure}[t]
			\centering
			\subfigure[MOS raw score distribution]{\includegraphics[width = 0.50\textwidth]{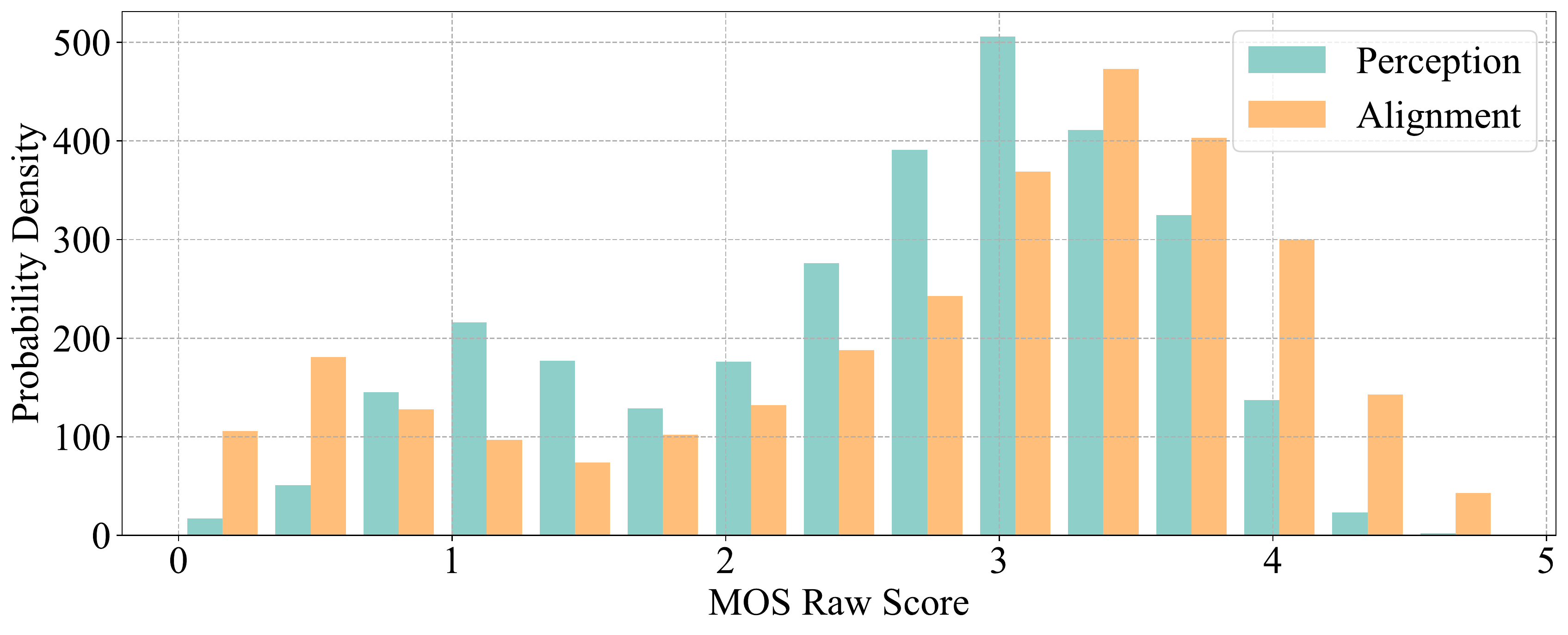}
			}
			\subfigure[MOS z-score distribution]{\includegraphics[width = 0.50\textwidth]{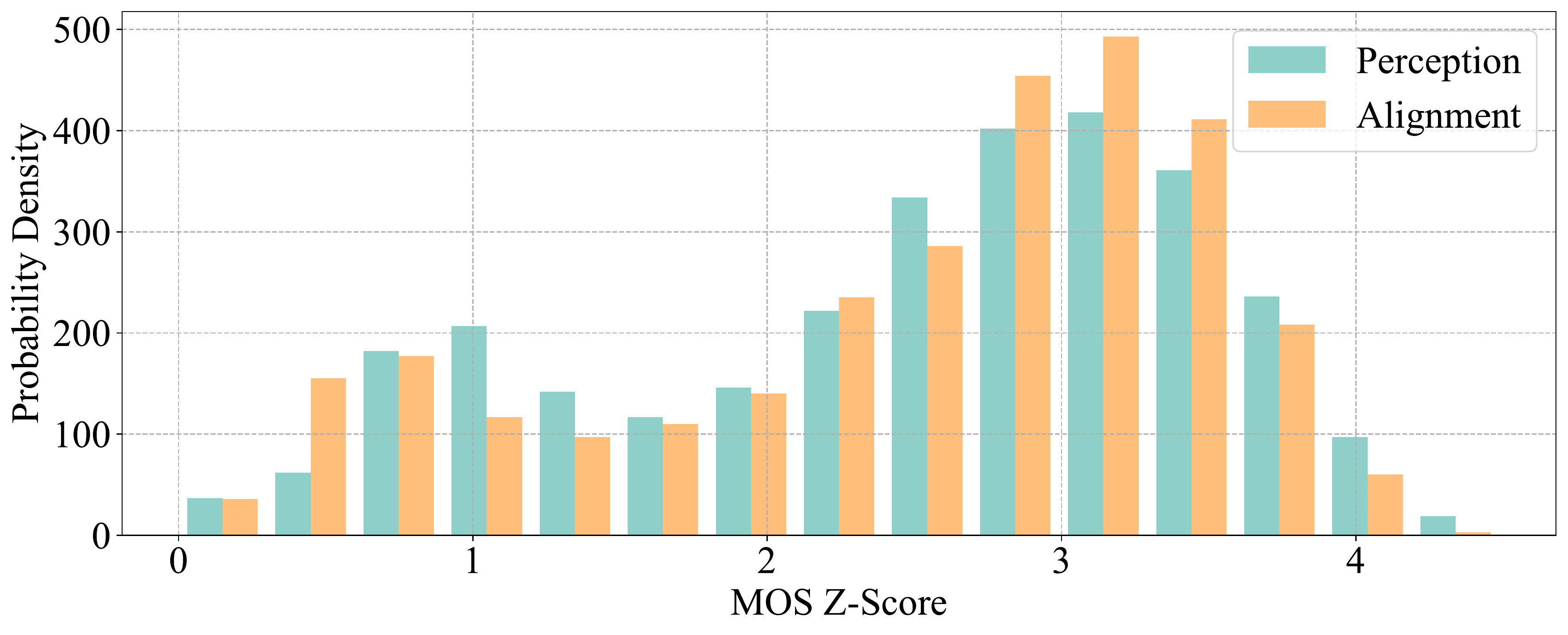}
			}
			\caption{Illustration of the raw/Z-score MOS probability distribution. The low/high-quality images covered at the perception and alignment levels are relatively even and abundant.}
			\label{fig:mos}
		\end{figure}

		The interface in Fig. \ref{fig:gui} allows viewers to browse the previous and next AGIs and move sliders ranging from 0 to 5 with a minimum interval of 0.1 as the quality score. A total of 21 graduate students (10 males and 11 females with 6 nationalities) participate in the experiment for 14 sessions. In case of visual fatigue, each session includes 213 images that limit the experiment time to half an hour. After collecting 2$\times$21$\times$2,982=125,244 quality ratings, we conduct post-processing for the final MOS score. First, we calculate the score correlation within all sessions and remove the outliers as the previous quality database \cite{other:quality1-xumai, quality:sisblim}. 
		After that, we normalize the average score $s$ for between each session to avoid inter-session scoring differences as:
		\begin{equation}
		s_{ij}(g) = r_{ij}(g) - \frac{1}{M}\sum_{i=g\cdot M}^{g\cdot M -1} r_{ij} + 2.5
		\end{equation}
		where $(i,j$) is the index of the image and viewer, $r$ stands for raw score and each session $g \in (0,13)$ contains $M$ images.
		Then subjective scores are converted to Z-scores $z_{ij}$ with the following formula:
		\begin{equation}
		z_{ij}=\frac{s_{ij}-\mu_j}{\sigma_j},
		\end{equation}
		where $\mu_j=\frac{1}{N}\sum_{i=0}^{N-1} r_{ij}$, $\sigma_j=\sqrt{\frac{1}{N-1}\sum_{i=0}^{N-1}(r_{ij}-\mu_i)^2}$ and $N$ is the number of viewers. Finally the MOS of image $j$ is computed by averaging the rescaled z-scores:
		
		\begin{equation}
		MOS_j=\frac{1}{N}\sum_{i=0}^{N-1} {\rm Res}(z_{ij}),
		\end{equation}
		where ${\rm Res}(\cdot)$ is the rescaling function. The MOS distribution in Fig. \ref{fig:mos} is consistent with previous works \cite{other:mos1} \cite{other:mos2} about subjective diversity.
		
		\subsection{Subjective Data Analysis}
		\label{sec:analysis}
		
		Although a large number of T2I AGI models ~\cite{gen:TextGAN, gen:StackGAN,gen:AttnGAN,gen:Cogview,gen:DALLE,gen:Parti,gen:GLIDE,gen:SD,gen:XL} have been developed in recent years, there is limited work \cite{review:sdmjde} investigating their generative performance. Under multiple models and different inputs, the quality (in terms of perception and alignment) of AGIs generated by the model is still an open question. Thanks to the abundant subjective quality scores and diverse prompts in the AGIQA-3K database, we conduct an in-depth analysis of this issue, and summarize the influencing factors of AGI subjective quality as follows:
		
		\textbf{AGI model}: The AGI model applied in the T2I generation task plays a major role in generating quality. With the same input prompt, the generation quality of different AGI models varies greatly.
		
		\textbf{Prompt length}: When the prompt is short, the model is easy to generate high-quality images; but when it reaches a certain length, it is difficult for AGI to meet the requirement of the entire prompt at the alignment level; even if successfully aligned, a certain level of perception will be sacrificed as a trade-off.
		
		\textbf{Prompt style}: The `style' item in the prompt is crucial to the generation quality. Considering the training process of the AGI model, the artistic style contained in the training data determines the generation quality of the AGI model for a certain style, which is reflected in perception and alignment together.
		
		\textbf{Model parameter}: The internal parameters of the model can affect the quality of AGI profoundly. CFG represents the `importance' ratio between perception and alignment. The larger the CFG, the model will value the alignment of AGI and prompt higher, but emphasize less perception accordingly; The number of iterations can also affect AGI's quality as the model gives an intermediate result when the iteration is insufficient.
		
		\begin{figure}[t]
			\centering
			\subfigure[Perception]{\includegraphics[width = 0.50\textwidth]{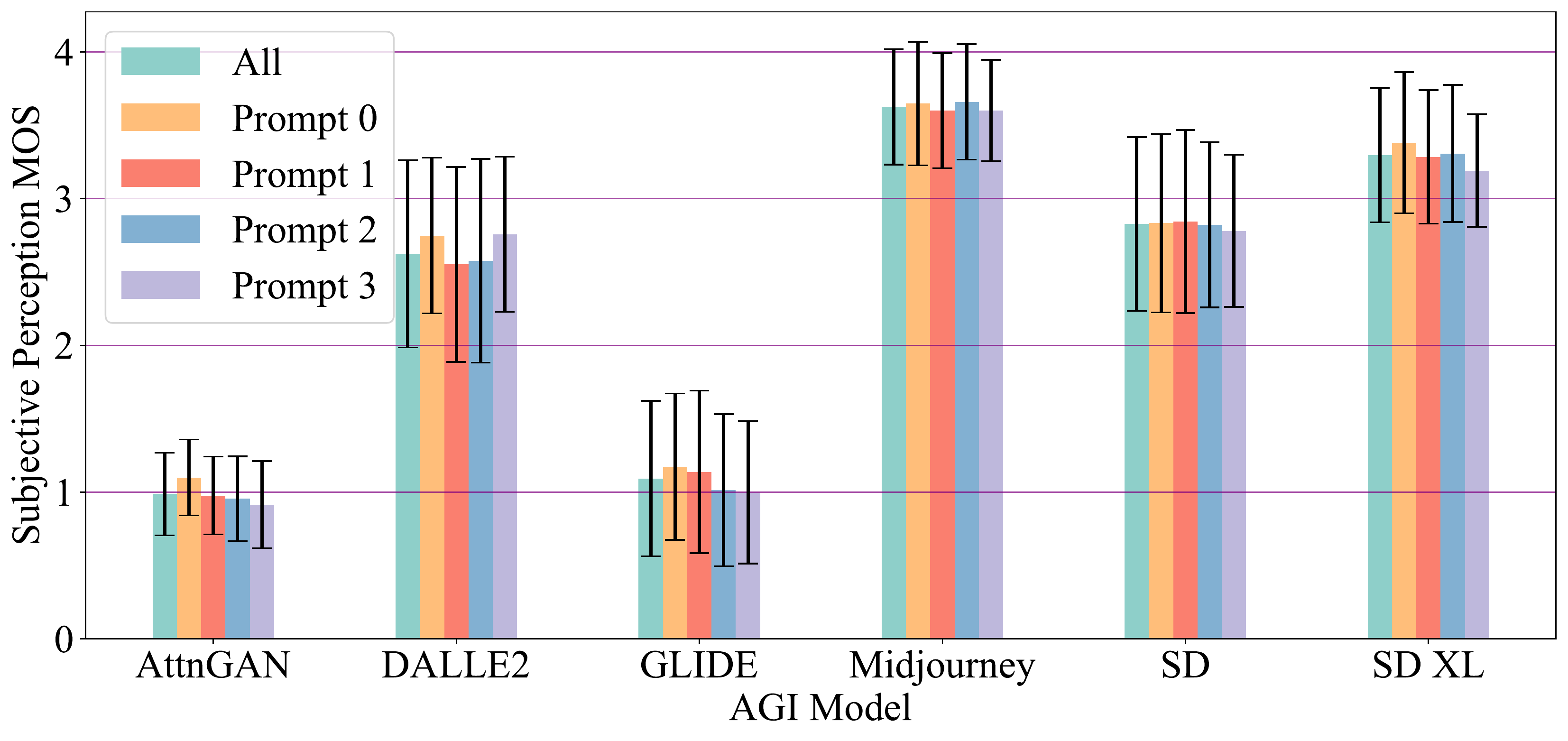}
			}
			\subfigure[Alignment]{\includegraphics[width = 0.50\textwidth]{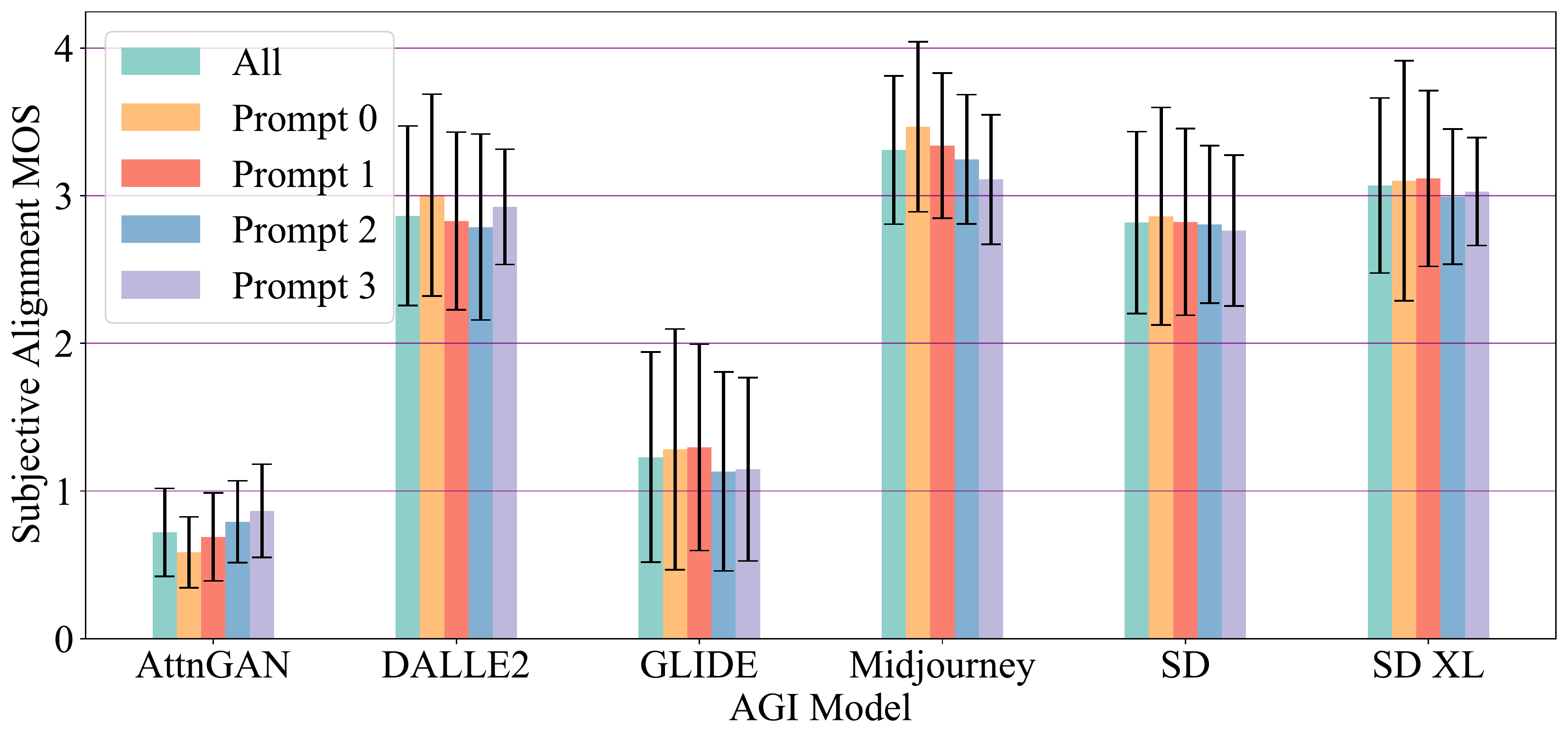}
			}
			\caption{The subjective quality of AGIs with different prompt lengths. The number after `prompt' is the amount of `Detail' and `Style' item. The data shows that as the prompt lengthens, the subjective quality of today's mainstream AGI models drop rapidly, especially the alignment score.}
			\label{fig:Prompt}
		\end{figure}
		
		Considering the above factors, we verified the subjective quality scores on prompt length and style under six AGI models, and showed the quality scores of Stable Diffusion and Midjourney after parameter tuning. The subjective quality under different prompt lengths (with only `Subject' as prompt 0, to `Subject' + 2$\times$`Detail' + `Style' as prompt 3) is shown in the Fig. \ref{fig:Prompt}, which reflects the generation quality of the six T2I AGI models is related to the prompt length, among which Midjourney and Stable Diffusion XL are the most representative. Midjourney's achieves satisfying perception quality regardless of the prompt length. However, the cost of such perception score is a decrease in alignment. The longer the prompt, the more items Midjourney will ignore, resulting in lower alignment. On the contrary, the alignment quality of Stable Diffusion XL is relatively stable, but it also shows a similar downward trend in perception quality.
		
		\begin{figure}[t]
			\centering
			\subfigure[Perception]{\includegraphics[width = 0.50\textwidth]{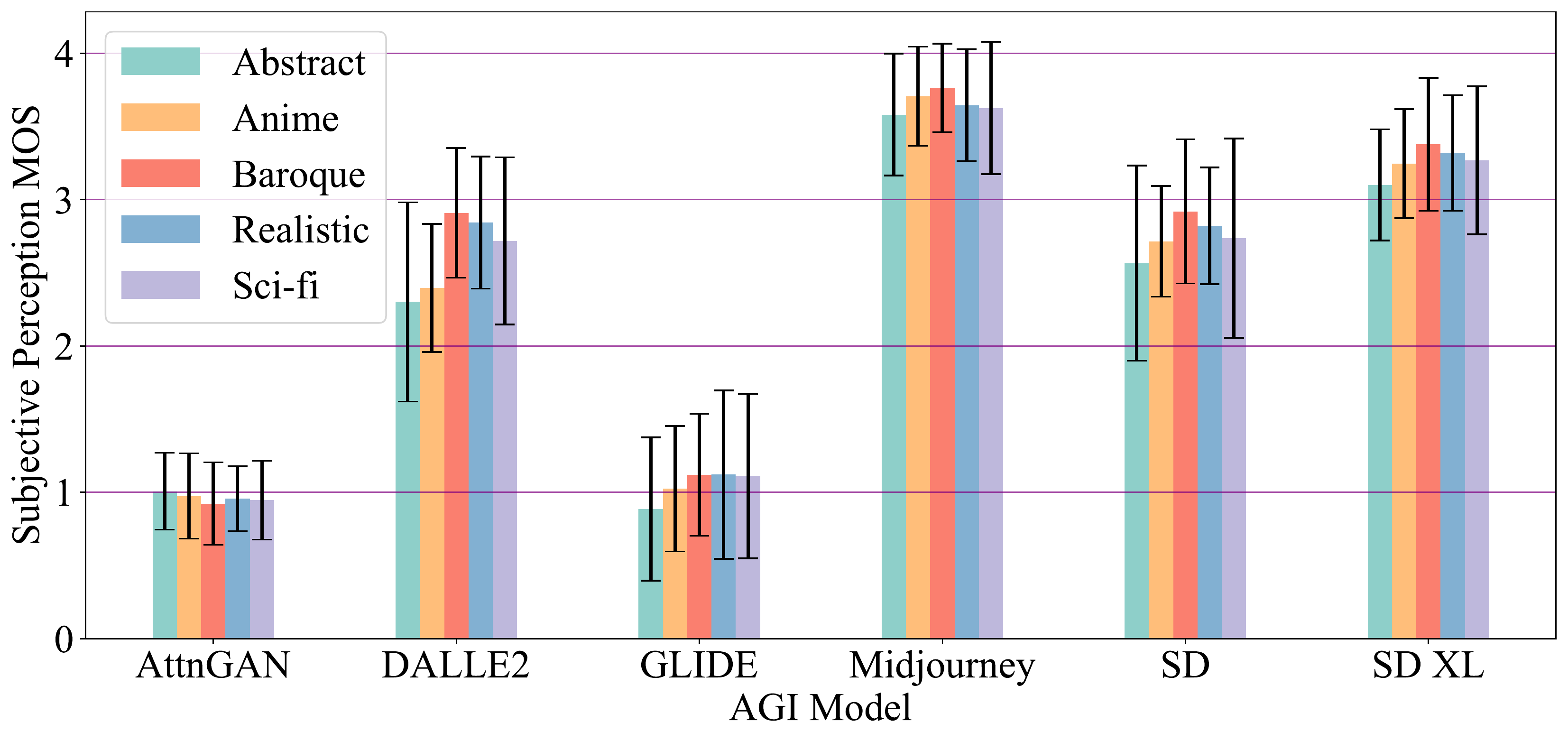}
			}
			\subfigure[Alignment]{\includegraphics[width = 0.50\textwidth]{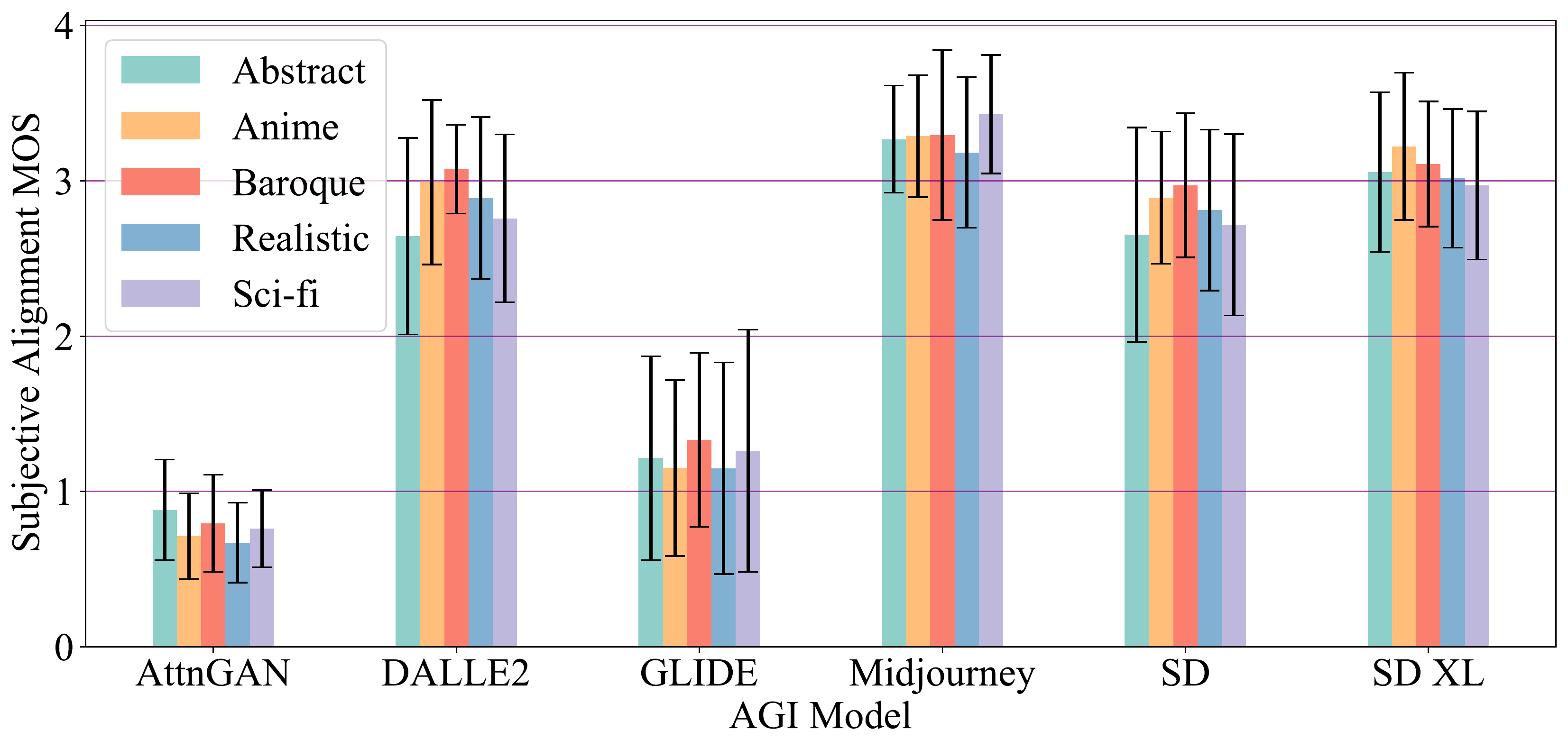}
			}
			\caption{The subjective quality of AGIs with different style. For both perception and alignment level, the quality of today's mainstream AGI models follows `Baroque'$>$`Anime\&Realistic'$>$`Abstract\&Sci-fi'.}
			\label{fig:Style}
		\end{figure}
		
		To analyze the performance of each T2I AGI model on different styles, we calculated the subjective quality scores of five styles: `Abstract', `Anime', `Baroque', `Realistic', and `Sci-fi' as shown in Fig. \ref{fig:Style}. Perception and alignment scores show that each model is good at generating `Baroque' style images, followed by `Anime' and `Realistic', and the worst performance on `Abstract' and `Sci-fi'. This is because the first three are relatively broad concepts, while the latter two are more specialized. Since the training data of the T2I AGI model usually contains a large number of NSIs, artworks and cartoon images, their generation result on the first three categories is fine; but for minority styles such as `Abstract' and `Sci-fi', the lack of training data  will lead to the defects of the final AGI in perception and alignment. Through horizontal comparison, we found that Midjourney has relatively good versatility in different styles, but the versatility of DALLE2 and Stable Diffusion still need to be improved. Especially for DALLE2, the difference in subjective perception scores between the two styles has reached 0.6. Therefore, for the future T2I AGI model, improving the versatility of different styles is under remarkable significance.
		
		\begin{figure}[t]
			\centering
			\subfigure[Stable Diffusion]{\includegraphics[height=40mm]{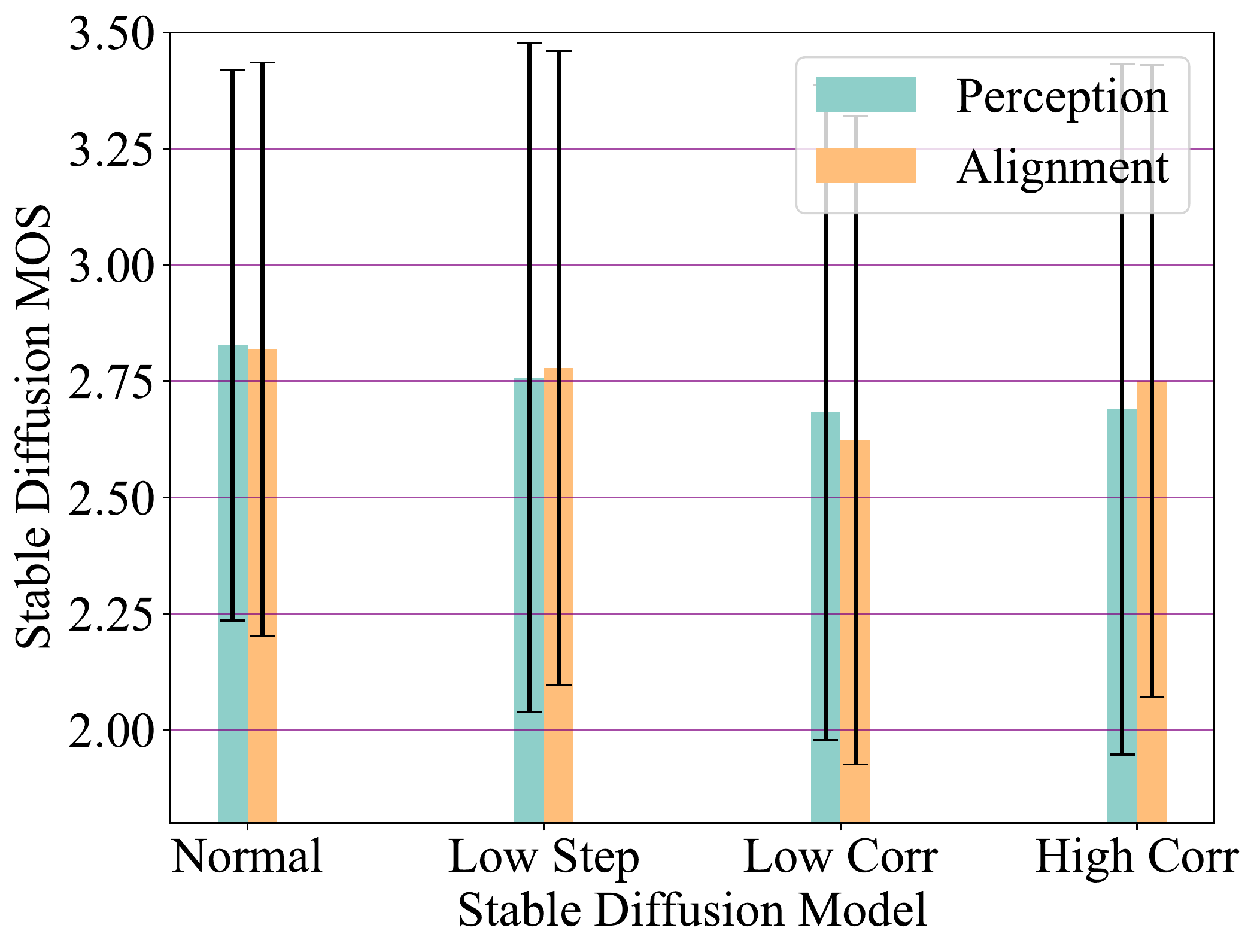}
			}
			\subfigure[Midjourney]{\includegraphics[height=40mm]{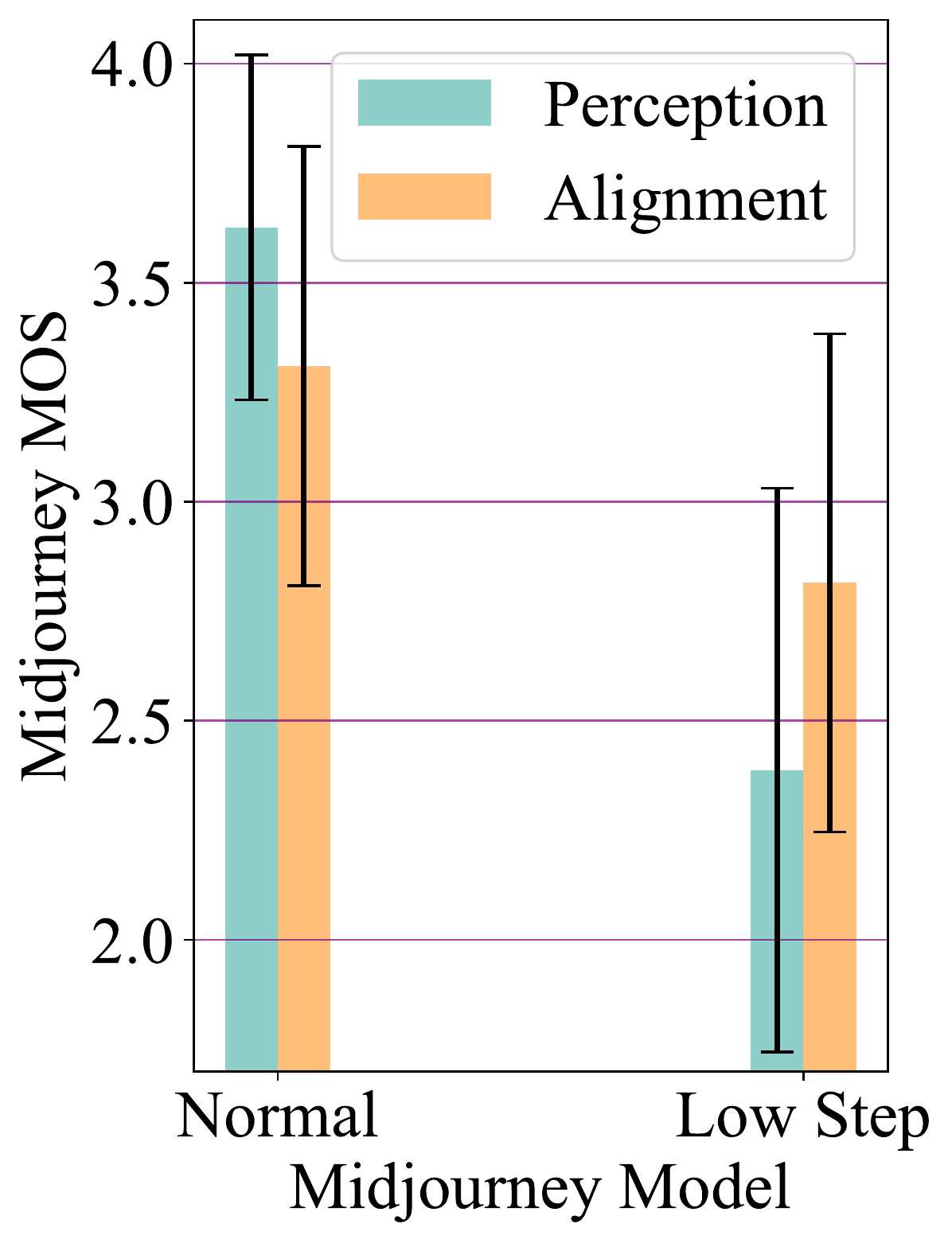}
			}
			\caption{The subjective quality of AGIs with different parameters. The result shows AGI models are sensitive to the number of iterations and CFG.}
			\label{fig:param}
		\end{figure}
		
		We also adjusted CFG and the number of iterations considering the impact of model internal parameters on AGI quality. Since Midjourney closed the field for CFG, this parameter is only adjusted to 0.5 times and 2 times the default value in Stable Diffusion; Meanwhile, we studied the subjective quality when the training step is insufficient by halving the number of iterations. The three situations above are characterized as `Low Corr' `High Corr' and `Low Step' in Fig. \ref{fig:param}. The data shows that Stable Diffusion has a strong tolerance for insufficient iterations, and both quality scores decline by less than 0.1; However, Midjourney’s quality drops significantly after the iteration was halved, especially since the perception score is almost reduced to the level of GLIDE. By observing the generation process of Midjourney, we find that in the first half of the step, the images only have blurred outlines, and certain details are rendered in the next half. So this kind of blur is likely to dominate the decline in perception score. For CFG, we found that increasing or decreasing this value will lead to a decrease in quality; if increased, the perception score will decrease significantly, and if decreased, the alignment will decrease more, which is consistent with the definition of CFG. It is worth mentioning that the decrease of one score will not increase the other, which shows the rationality of the default CFG value in Stable Diffusion, and it is not recommended to adjust it at will.
		
		\section{Alignment Quality Metric}
		
		\subsection{Framework}
		
\begin{figure*}[t]
			\centering
			\includegraphics[width=\textwidth]{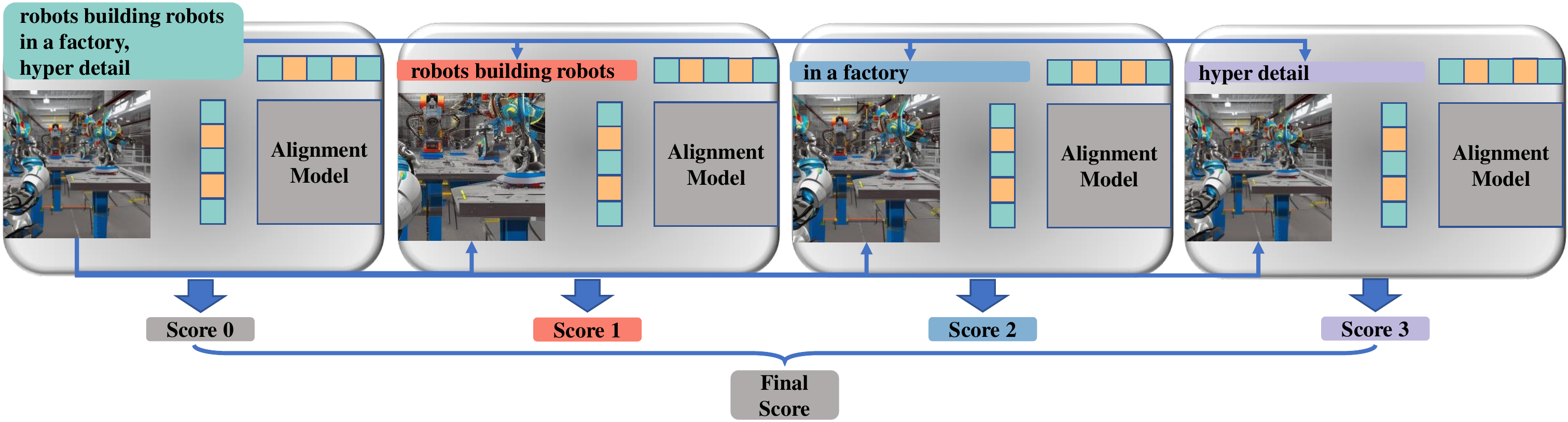}
			\caption{The framework of the StairReward alignment assessment model. The example prompt includes 3 morpheme.}
			\label{fig:framework}
		\end{figure*}

		Considering the remarkable variance of the subjective alignment quality score in Sec. \ref{sec:analysis}, we propose an objective alignment quality assessment metric StairReward. This method disassembles the alignment quality assessment to the morpheme level for the first time, instead of using the entire prompt as the previous method \cite{align:clip, database/align:HPS}. The framework of StairReward is shown in Fig. \ref{fig:framework}, which divides the prompt into multiple morphemes while cutting the whole picture into multiple stairs and gives the final score through their one-to-one alignment. The detail of each component in the proposed model is described as follows. 
		
		\subsection{Prompt segmentation}
		
		A prompt contains multiple morphemes, while a human has different saliency in its different position. For alignment quality, the earlier morphemes have a greater impact on the subjective score, while the later morphemes have less impact. Therefore, we first decompose the prompt into morphemes, so that the objective quality model is more consistent with the subjective perception mechanism. Considering that there are certain differences between prompt \cite{other:cut1,other:cut2} and natural language \cite{other:cut3}, it is not reasonable to directly use previous word segmentation algorithms \cite{other:seg} to split prompt. Therefore, We adopt our own prompt segmentation method. By observing a large number of prompts in DiffusionDB \cite{database:DiffusionDB}, we found that prepositions and punctuation marks are the two most common elements that separate prompts. Therefore, the morpheme sequence $({p_1},{p_2} \cdots {p_K})$ obtained by prompt ${p_0}$ is as follows:
		
		\begin{equation}
		({p_1},{p_2} \cdots {p_K}) = {\rm Split}(p_0)
		\end{equation}
		where $K$ represents the number of morphemes and ${\rm Split}(\cdot)$ is a prompt segmentation function based on prepositions and punctuation. Thus, we successfully break the prompt into several morphemes with different importance.

		\subsection{Image Cutting}
		
\begin{figure}[t]
			\centering
			\subfigure[Average CLIP score]{\includegraphics[width = 0.50\textwidth]{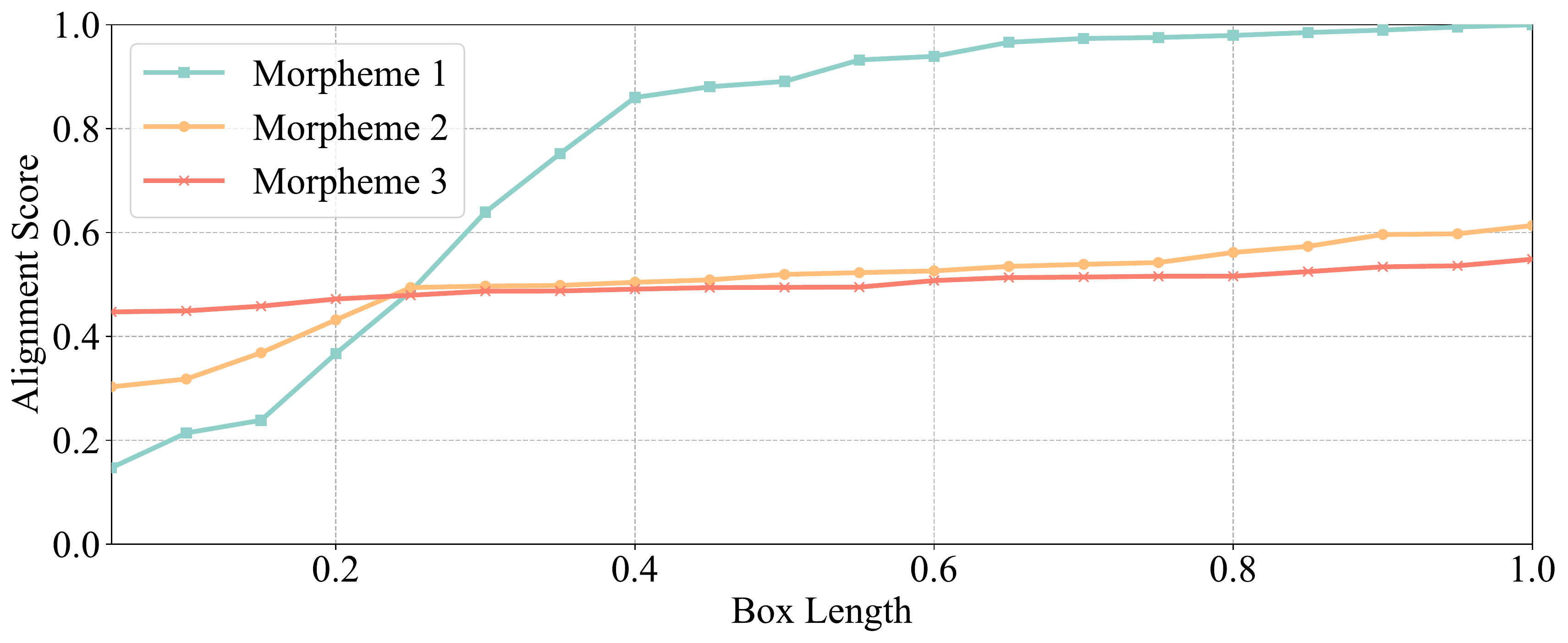}}
			\subfigure[SRoCC between CLIP and subjective alignment score]{\includegraphics[width = 0.50\textwidth]{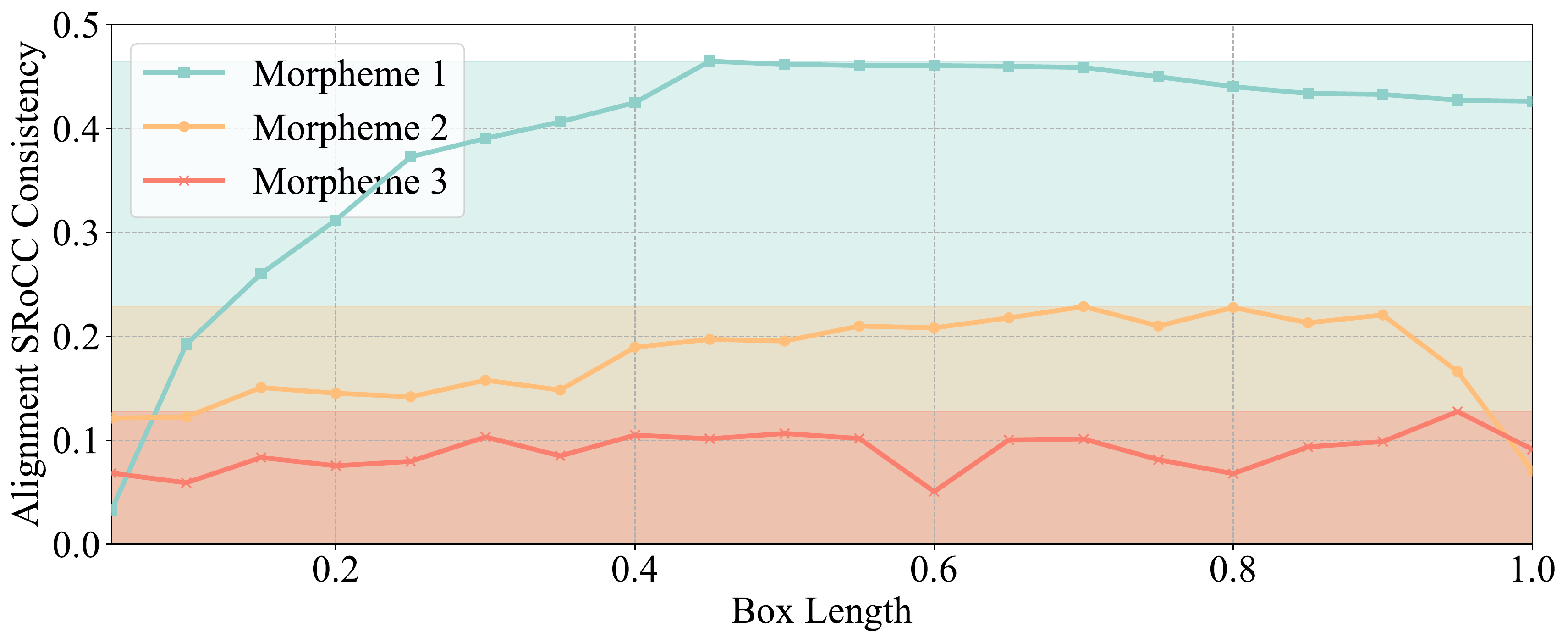}}
			\caption{The feature of different box lengths in AGIQA-3K with 3 morphemes as an example. (a) shows that when the box size reaches a certain level, a sub-graph related to morphemes can be successfully collected; (b) proves for the morpheme whose position is more preceding, the optimal box size is smaller.}
			\label{fig:stair}
		\end{figure}

		Since the prompt is split as a morpheme in AGIs, we need to first locate the corresponding region of the morpheme in AGI and then compute the T2I alignment score. To avoid extra complexity in analyzing image content, we assume that the center of the image contains the most information and the edges the least. Therefore, we use the same box as the center of the image for sampling and cut the image into different stairs by adjusting the length of the box. Fig. \ref{fig:stair} (a) proves the rationality of this method. When the prompt contains three morphemes, the growth of the clip score of the first morpheme and the picture suddenly slows down after the box length reaches 0.5, while the clip score of the third morpheme and the picture keeps steady growth. It can be seen that the first morpheme mainly corresponds to the central part of the picture, and the later morphemes correspond to a larger area. Therefore, based on the morpheme number, the stair-image $S_k$ is set as:
		
		\begin{equation}
		{I_k} = {\rm Box}_{L=\frac{1}{2} + \frac{k - 1}{2(K - 1)}}({I_0})
		\end{equation}
		where $k\in [1,K]$ is the index of morphemes and $L$ is the box length cutting original image $I_0$. Fig. \ref{fig:stair} (b) shows Spearman Rank-order Correlation Coefficien (SRoCC) turning points in 0.5/0.75/1 box length for 3 morphemes, which proves that selecting stair-images as described above can enhance the consistency of objective and subjective scores.
		
		\subsection{Final Score Combination}
		
		With sub-images and morphemes, we compute their alignment scores one by one. As ImageReward \cite{database/align:ImageReward} uses extensive training data, it is suitable for prompts of different lengths and fits the morpheme. Thus, we choose ImageReward as our alignment model. After calculating the above scores, since each score has a different impact on the overall alignment, we set the weight of the latter score to half of the previous one referring to the vertical axis in Fig. \ref{fig:stair} (b). Finally, to reserve information between morphemes, we also calculated the alignment score between the entire image and the prompt, and then added the above scores for the final score $F$:
		
		\begin{equation}
		F = {\rm A}({p_0},{I_0}) + \sum\limits_{k = 1}^K {\frac{{\rm A}({p_k},{I_k})}{2^k}} /(1 - \frac{1}{2^K})
		\end{equation}
		where ${\rm A} (\cdot)$ is the alignment function. Thus, we split the prompt into several morphemes and separate the image into different stairs. Through their one-to-one correspondence, an effective alignment score is given.

		\section{Experiment Results}
		
		\subsection{Experiment Settings}
		
		To benchmark the performance of AGI perception and alignment metrics, three commonly used indicators, including SRoCC, Kendall Rank-order Correlation Coefficient (KRoCC), and  Pearson Linear Correlation Coefficient (PLCC) are applied to evaluate the consistency between the predicted score and the subjective MOS, among which the SRoCC and KRoCC indicate the prediction monotonicity while the PLCC represents the prediction accuracy. To map the predicted scores to MOSs, a five-parameter logistic function is applied, which is a standard practice suggested in \cite{other:stat}:
		\begin{equation}
		\hat{X}=\alpha_{1}\left(0.5-\frac{1}{1+e^{\alpha_{2}\left(X-\alpha_{3}\right)}}\right)+\alpha_{4} X+\alpha_{5},
		\end{equation}
		where $\{\alpha_{i} \mid i=1,2, \ldots, 5\}$ represent the parameters for fitting, $y$ and $\hat{y}$ stand for predicted and fitted scores respectively.
		
		We select a wide range of AGI perception and alignment benchmarks for comparison. For perception, only No-Reference (NR) metrics are selected considering the absence of reference in the T2I AGI task as Sec. \ref{sec:quality} reviewed:
		
		\begin{itemize}
			\item Handcrafted-based models: This group includes four mainstream perceptual quality metrics, namely CEIQ \cite{quality:CEIQ}, DSIQA \cite{quality:DSIQA}, NIQE \cite{quality:NIQE}, and Sisblim \cite{quality:sisblim}. These models extract handcrafted features based on prior knowledge about image quality.
			\item Loss-function models: This group includes three loss-function that are commonly used in AGI iterations, namely FID \cite{quality:fid}, InCeption Score (ICS) \cite{quality:ics} and KID \cite{quality:kid}. The FID and KID measure the distance between AGIs and the MS-COCO \cite{database:mscoco} database. 
			
			\item SVR-based models: This group includes BMPRI \cite{quality:BMPRI}, GMLF \cite{quality:GMLF}, HIGRADE \cite{quality:Higrade}. These models combine hand-crafted features from a Support Vector Regression (SVR) to represent perceptual quality.
			
			\item DL-based models: This group includes the latest deep learning (DL) metrics, namely DBCNN \cite{quality:DBCNN}, CLIPIQA \cite{quality:CLIPIQA}, CNNIQA \cite{quality:CNNIQA}, HyperNet \cite{quality:HyperNet}. These models characterize quality-aware information by training deep neural networks from labeled data. 
		\end{itemize}
		For alignment, we select the most popular CLIP \cite{align:clip} model and the latest ImageReward \cite{database/align:ImageReward}, HPS \cite{database/align:HPS}, PickScore \cite{database/align:PickAPic}, and the StairScore we proposed.
		
		The AGIQA-3K is split randomly in an 80/20 ratio for training/testing while ensuring the image with the same object label falls into the same set. The partitioning and evaluation process is repeated several times for a fair comparison while considering the computational complexity, and the average result is reported as the final performance. For SVR models, the repeating time is 1,000, implemented by LIBSVM \cite{other:libsvm} with radial basis function (RBF) kernel. For DL models, we use the pyiqa \cite{other:pyiqa} framework with 10 similar repeatings. The Adam optimizer \cite{other:adam} (with an initial learning rate of 0.00001 and batch size 40) is used for 100-epochs training on an NVIDIA GTX 4090Ti GPU.
		
		\begin{table*}[tbph]
			\caption{Perception metric performance results on the AGIQA-3K database and different subsets from different T2I AGI models. The best performance results are marked in {\bf\textcolor{red}{RED}} and the second performance results are marked in {\bf\textcolor{blue}{BLUE}}.}
			\label{tab:perception}
			\subtable[All AGIQA-3K database and three different subsets from different T2I AGI model groups]{
				\begin{tabular}{l|c|ccc|ccc|ccc|ccc}
					\toprule
					\multicolumn{1}{c|}{\multirow{2}{*}{Type}} & \multirow{2}{*}{Metric} & \multicolumn{3}{c|}{All} & \multicolumn{3}{c|}{Bad Model} & \multicolumn{3}{c|}{Medium Model} & \multicolumn{3}{c}{Good Model} \\ \cline{3-14} 
					&                         & SRoCC  & KRoCC  & PLCC   & SRoCC    & KRoCC    & PLCC     & SRoCC     & KRoCC     & PLCC      & SRoCC    & KRoCC    & PLCC     \\ \hline
					\multirow{4}{*}{\begin{tabular}[c]{@{}l@{}}Hand\\ crafted-\\ based\end{tabular}}    & CEIQ\cite{quality:CEIQ}                    & 0.3228 & 0.2220 & 0.4166 & 0.1754   & 0.1141   & 0.2094   & 0.2775    & 0.1868    & 0.3043    & 0.1743   & 0.1161   & 0.1643   \\
					& DSIQA\cite{quality:DSIQA}                   & 0.4955 & 0.3403 & 0.5488 & 0.1908   & 0.1331   & 0.3139   & 0.2140    & 0.1469    & 0.3655    & 0.1665   & 0.1120   & 0.2520   \\
					& NIQE\cite{quality:NIQE}                    & 0.5623 & 0.3876 & 0.5171 & 0.2031   & 0.1354   & 0.3309   & 0.2259    & 0.1483    & 0.2526    & 0.1750   & 0.1172   & 0.2533   \\
					& Sisblim\cite{quality:sisblim}                 & 0.5479 & 0.3788 & 0.6477 & 0.2887   & 0.2012   & 0.3341   & 0.0540    & 0.0357    & 0.2932    & 0.0417   & 0.0266   & 0.2110   \\ \hline
					\multirow{3}{*}{\begin{tabular}[c]{@{}l@{}}Loss-\\ function\end{tabular}} & FID\cite{quality:fid}                     & 0.1733 & 0.1158 & 0.1860 & 0.1836   & 0.1249   & 0.1938   & 0.1402    & 0.0929    & 0.1614    & 0.0562   & 0.0348   & 0.0798   \\
					& ICS\cite{quality:ics}                     & 0.0931 & 0.0626 & 0.0964 & 0.0243   & 0.0179   & 0.1692   & 0.0797    & 0.0534    & 0.1693    & 0.0856   & 0.0574   & 0.1042   \\
					& KID\cite{quality:kid}                     & 0.1023 & 0.0692 & 0.0786 & 0.0028   & 0.0077   & 0.0187   & 0.1279    & 0.0839    & 0.0860    & 0.0704   & 0.0472   & 0.0614   \\ \hline
					\multirow{3}{*}{\begin{tabular}[c]{@{}l@{}}SVR-\\ based\end{tabular}}  & BMPRI\cite{quality:BMPRI}                   & 0.6794 & 0.4976 & 0.7912 & 0.3686   & 0.2583   & 0.4076   & 0.2374    & 0.1650    & 0.3760    & 0.2046   & 0.1385   & 0.2212   \\
					& GMLF\cite{quality:GMLF}                    & 0.6987 & 0.5119 & 0.8181 & 0.3942   & 0.2774   & 0.4798   & 0.2578    & 0.1751    & 0.4036    & 0.0018   & 0.0023   & 0.0834   \\
					& Higrade\cite{quality:Higrade}                 & 0.6171 & 0.4410 & 0.7056 & 0.3017   & 0.2106   & 0.3001   & 0.2376    & 0.1619    & 0.2861    & 0.2020   & 0.1398   & 0.2164   \\ \hline
					\multirow{4}{*}{\begin{tabular}[c]{@{}l@{}}DL-\\ based\end{tabular}}    & DBCNN\cite{quality:DBCNN}                   & 0.8207 & 0.6336 & {\bf\textcolor{blue}{0.8759}} & {\bf\textcolor{red}{0.5520}}   & {\bf\textcolor{red}{0.3958}}   & {\bf\textcolor{red}{0.6825}}   & {\bf\textcolor{blue}{0.5011}}    & {\bf\textcolor{blue}{0.3531}}    & {\bf\textcolor{blue}{0.5575}}    & 0.4288   & 0.2975   & 0.4853   \\
					& CLIPIQA\cite{quality:CLIPIQA}                & {\bf\textcolor{red}{0.8426}} & {\bf\textcolor{blue}{0.6468}} & 0.8053 & 0.1882   & 0.1255   & 0.2549   & {\bf\textcolor{red}{0.6537}}    & {\bf\textcolor{red}{0.4693}}    & {\bf\textcolor{red}{0.6014}}    & {\bf\textcolor{blue}{0.5038}}   & {\bf\textcolor{blue}{0.3407}}   & {\bf\textcolor{blue}{0.5081}}   \\
					& CNNIQA\cite{quality:CNNIQA}                  & 0.7478 & 0.5580 & 0.8469 & 0.3233   & 0.2275   & 0.4547   & 0.4278    & 0.2807    & 0.4534    & 0.3952   & 0.2805   & 0.4517   \\
					& HyperNet\cite{quality:HyperNet}                & {\bf\textcolor{blue}{0.8355}} & {\bf\textcolor{red}{0.6488}} & {\bf\textcolor{red}{0.8903}} & {\bf\textcolor{blue}{0.5086}}   & {\bf\textcolor{blue}{0.3628}}   & {\bf\textcolor{blue}{0.5985}}   & 0.4687    & 0.3260    & 0.5480    & {\bf\textcolor{red}{0.5562}}   & {\bf\textcolor{red}{0.3927}}   & {\bf\textcolor{red}{0.6149}}   \\ 
					\bottomrule
			\end{tabular}}

			\subtable[Four different AGIQA-3K subsets with different prompt length]{
				\begin{tabular}{l|c|ccc|ccc|ccc|ccc}
					\toprule
					\multicolumn{1}{c|}{\multirow{2}{*}{Type}}      & \multirow{2}{*}{Metric} & \multicolumn{3}{c|}{Prompt 0} & \multicolumn{3}{c|}{Prompt 1} & \multicolumn{3}{c|}{Prompt 2} & \multicolumn{3}{c}{Prompt 3} \\ \cline{3-14} 
					&                         & SRoCC    & KRoCC    & PLCC    & SRoCC    & KRoCC    & PLCC    & SRoCC    & KRoCC    & PLCC    & SRoCC    & KRoCC   & PLCC    \\ \hline
					\multirow{4}{*}{\begin{tabular}[c]{@{}l@{}}Hand\\ crafted-\\ based\end{tabular}} & CEIQ\cite{quality:CEIQ}                    & 0.3238   & 0.2217   & 0.4080  & 0.3175   & 0.2186   & 0.3972  & 0.3415   & 0.2349   & 0.4396  & 0.2811   & 0.1931  & 0.4338  \\
					& DSIQA\cite{quality:DSIQA}                   & 0.5110   & 0.3522   & 0.5419  & 0.5000   & 0.3448   & 0.5631  & 0.4651   & 0.3195   & 0.5298  & 0.5221   & 0.3599  & 0.5743  \\
					& NIQE\cite{quality:NIQE}                    & 0.5975  & 0.4104  & 0.5350 & 0.5478  & 0.3754  & 0.5138 & 0.5471  & 0.3762  & 0.4909 & 0.5669  & 0.4015 & 0.5819 \\
					& Sisblim\cite{quality:sisblim}                 & 0.5629  & 0.3883  & 0.6573 & 0.5343  & 0.3710  & 0.6439 & 0.5518  & 0.3811  & 0.6472 & 0.5532  & 0.3825 & 0.6530 \\ \hline
					\multirow{3}{*}{\begin{tabular}[c]{@{}l@{}}Loss-\\ function\end{tabular}}      & FID\cite{quality:fid}                     & 0.1590  & 0.1078  & 0.1629 & 0.2113  & 0.1417  & 0.2248 & 0.1449  & 0.0967  & 0.1577 & 0.1389  & 0.0932 & 0.1751 \\
					& ICS\cite{quality:ics}                     & 0.0922  & 0.0612  & 0.1289 & 0.1396  & 0.0943  & 0.1155 & 0.0422  & 0.0283  & 0.0486 & 0.0693  & 0.0461 & 0.1514 \\
					& KID\cite{quality:kid}                     & 0.1228   & 0.0837   & 0.1031  & 0.1255   & 0.0844   & 0.1010  & 0.0473   & 0.0337   & 0.0215  & 0.1120   & 0.0763  & 0.0908  \\ \hline
					\multirow{3}{*}{\begin{tabular}[c]{@{}l@{}}SVR-\\ based\end{tabular}}       & BMPRI\cite{quality:BMPRI}                   & 0.6880   & 0.5086   & 0.7980  & 0.6888   & 0.5055   & 0.7918  & 0.6919   & 0.5123   & 0.8050  & 0.6661   & 0.4892  & 0.8104  \\
					& GMLF\cite{quality:GMLF}                    & 0.7045   & 0.5160   & 0.8222  & 0.7081   & 0.5220   & 0.8211  & 0.7051   & 0.5212   & 0.8229  & 0.6821   & 0.5047  & 0.8377  \\
					& Higrade\cite{quality:Higrade}                 & 0.6578   & 0.4781   & 0.7308  & 0.6133   & 0.4396   & 0.6986  & 0.6360   & 0.4568   & 0.7246  & 0.5964   & 0.4248  & 0.6872  \\ \hline
					\multirow{4}{*}{\begin{tabular}[c]{@{}l@{}}DL-\\ based\end{tabular}}         & DBCNN\cite{quality:DBCNN}                   & 0.8162   & 0.6271   & {\bf\textcolor{blue}{0.8998}}  & {\bf\textcolor{blue}{0.8246}}   & {\bf\textcolor{blue}{0.6380}}   & {\bf\textcolor{blue}{0.8657}}  & 0.8160   & 0.6340   & {\bf\textcolor{blue}{0.8562}}  & 0.8051   & 0.6166  & 0.8846  \\
					& CLIPIQA\cite{quality:CLIPIQA}                & {\bf\textcolor{blue}{0.8458}}   & {\bf\textcolor{blue}{0.6541}}   & 0.8059  & {\bf\textcolor{red}{0.8471}}   & {\bf\textcolor{red}{0.6492}}   & 0.8065  & {\bf\textcolor{blue}{0.8365}}   & {\bf\textcolor{blue}{0.6364}}   & 0.7734  & {\bf\textcolor{red}{0.8364}}   & {\bf\textcolor{blue}{0.6407}}  & 0.8143  \\
					& CNNIQA\cite{quality:CNNIQA}                  & 0.7992   & 0.6127   & 0.8851  & 0.7082   & 0.5159   & 0.8104  & 0.7780   & 0.5868   & 0.8537  & 0.7620   & 0.5839  & {\bf\textcolor{blue}{0.8966}}  \\
					& HyperNet\cite{quality:HyperNet}                & {\bf\textcolor{red}{0.8641}}   & {\bf\textcolor{red}{0.6865}}   & {\bf\textcolor{red}{0.9238}}  & 0.8214   & 0.6320   & {\bf\textcolor{red}{0.8688}}  & {\bf\textcolor{red}{0.8376}}   & {\bf\textcolor{red}{0.6642}}   & {\bf\textcolor{red}{0.8942}}  & {\bf\textcolor{blue}{0.8237}}   & {\bf\textcolor{red}{0.6457}}  & {\bf\textcolor{red}{0.9083}}  \\ \bottomrule
			\end{tabular}}
			
			\subtable[Four different AGIQA-3K subsets with different style]{
				\begin{tabular}{l|c|ccc|ccc|ccc|ccc}
					\toprule
					\multicolumn{1}{c|}{\multirow{2}{*}{Type}}      & \multirow{2}{*}{Metric} & \multicolumn{3}{c|}{Abstract \& Sci-fi Style} & \multicolumn{3}{c|}{Anime \& Realistic Style} & \multicolumn{3}{c|}{Baroque Style} & \multicolumn{3}{c}{No Style} \\ \cline{3-14} 
					&                         & SRoCC    & KRoCC    & PLCC    & SRoCC    & KRoCC    & PLCC    & SRoCC    & KRoCC    & PLCC    & SRoCC    & KRoCC   & PLCC    \\ \hline
					\multirow{4}{*}{\begin{tabular}[c]{@{}l@{}}Hand\\ crafted-\\ based\end{tabular}} & CEIQ\cite{quality:CEIQ}                    & 0.3185 & 0.2193 & 0.3990 & 0.2795   & 0.1915   & 0.4201   & 0.2562    & 0.1752    & 0.3845    & 0.3689   & 0.2557   & 0.4470   \\
					& DSIQA\cite{quality:DSIQA}                   & 0.5546 & 0.3847 & 0.5238 & 0.5286   & 0.3726   & 0.6226   & 0.6020    & 0.4276    & 0.5994    & 0.4908   & 0.3339   & 0.5565   \\
					& NIQE\cite{quality:NIQE}                    & 0.4887 & 0.3393 & 0.4286 & 0.5015   & 0.3493   & 0.5275   & 0.5759    & 0.3960    & 0.5290    & 0.5748   & 0.3944   & 0.5365   \\
					& Sisblim\cite{quality:sisblim}                 & 0.5819 & 0.4115 & 0.6293 & 0.5360   & 0.3707   & 0.6712   & 0.5498    & 0.3792    & 0.6885    & 0.5651   & 0.3921   & 0.6524   \\ \hline
					\multirow{3}{*}{\begin{tabular}[c]{@{}l@{}}Loss-\\ function\end{tabular}} & FID\cite{quality:fid}                     & 0.1935 & 0.1297 & 0.2362 & 0.1347   & 0.0901   & 0.1107   & 0.1504    & 0.0970    & 0.1628    & 0.1663   & 0.1107   & 0.1786   \\
					& ICS\cite{quality:ics}                     & 0.0963 & 0.0648 & 0.1563 & 0.0873   & 0.0601   & 0.1536   & 0.0909    & 0.0584    & 0.0305    & 0.0789   & 0.0528   & 0.0802   \\
					& KID\cite{quality:kid}                     & 0.0360 & 0.0232 & 0.0153 & 0.0423   & 0.0268   & 0.0645   & 0.1889    & 0.1258    & 0.2156    & 0.1089   & 0.0739   & 0.0703   \\ \hline
					\multirow{3}{*}{\begin{tabular}[c]{@{}l@{}}SVR-\\ based\end{tabular}}  & BMPRI\cite{quality:BMPRI}                   & 0.6930 & 0.5144 & 0.7875 & 0.6247   & 0.4628   & 0.7811   & 0.6169    & 0.4439    & 0.7932    & 0.7061   & 0.5188   & 0.8057   \\
					& GMLF\cite{quality:GMLF}                    & 0.7573 & 0.5763 & 0.8233 & 0.6506   & 0.4839   & 0.8011   & 0.6777    & 0.5005    & 0.8534    & 0.7003   & 0.5128   & 0.8223   \\
					& Higrade\cite{quality:Higrade}                 & 0.6034 & 0.4333 & 0.6793 & 0.5729   & 0.4170   & 0.6943   & 0.5858    & 0.4280    & 0.7191    & 0.6240   & 0.4462   & 0.6964   \\ \hline
					\multirow{4}{*}{\begin{tabular}[c]{@{}l@{}}DL-\\ based\end{tabular}}   & DBCNN\cite{quality:DBCNN}                   & 0.8132 & 0.6240 & 0.8523 & {\bf\textcolor{blue}{0.8312}}   & {\bf\textcolor{blue}{0.6487}}   & {\bf\textcolor{red}{0.8774}}   & {\bf\textcolor{red}{0.7794}}    & {\bf\textcolor{red}{0.5975}}    & {\bf\textcolor{red}{0.8901}}    & 0.8067   & 0.6216   & {\bf\textcolor{blue}{0.8748}}   \\
					& CLIPIQA\cite{quality:CLIPIQA}                & {\bf\textcolor{red}{0.8968}} & {\bf\textcolor{red}{0.7037}} & 0.8207 & {\bf\textcolor{red}{0.8678}}   & {\bf\textcolor{red}{0.6753}}   & 0.8187   & {\bf\textcolor{blue}{0.7750}}    & {\bf\textcolor{blue}{0.5919}}    & 0.8222    & {\bf\textcolor{blue}{0.8313}}   & {\bf\textcolor{blue}{0.6371}}   & 0.7898   \\
					& CNNIQA\cite{quality:CNNIQA}                  & 0.7729 & 0.6115 & {\bf\textcolor{blue}{0.8544}} & 0.7581   & 0.5749   & 0.8295   & 0.6845    & 0.4885    & 0.8597    & 0.7380   & 0.5466   & 0.8481   \\
					& HyperNet\cite{quality:HyperNet}                & {\bf\textcolor{blue}{0.8303}} & {\bf\textcolor{blue}{0.6618}} & {\bf\textcolor{red}{0.8771}} & 0.8060   & 0.6245   & {\bf\textcolor{blue}{0.8745}}   & 0.7676    & 0.5807    & {\bf\textcolor{blue}{0.8646}}    & {\bf\textcolor{red}{0.8317}}   & {\bf\textcolor{red}{0.6470}}   & {\bf\textcolor{red}{0.8899}}  \\ \bottomrule
			\end{tabular}}
		\end{table*}
		
		\subsection{Experiment Results and Discussion}
		
		Tab \ref{tab:perception} lists the performance result of different perception models on the proposed AGIQA-3K database. To analyze the assessment consistency of the perception model and subjective score generated by different T2I AGI models, we divide six AGI models into three groups, namely bad model (AttnGAN\cite{gen:AttnGAN}, GLIDE\cite{gen:GLIDE}), medium model (DALLE2\cite{gen:DALLE2}, Stable Diffusion\cite{gen:SD}), and good model (Midjourney\cite{gen:MJ}, Stable Diffusion XL) based on the subjective performance/alignment score in Fig. \ref{fig:Prompt}. Tab. \ref{tab:perception} (a) shows that comparing with loss-function in AGI iterations, the other three types of models are more compatible with HVS, especially the DL model with an overall SRoCC about 0.8. However, their performance is not that satisfying on three subsets generated by different AGI models. For each subset, even the SRoCC of the best perception model can only reach about 0.5. There are also significant differences in the analytical capabilities of different models for low/high-quality content. For example, the overall performance of CLIPIQA and DBCNN is comparable, but CLIP shows a significant advantage in analyzing aesthetic features in the good model subset, and the performance is not satisfactory when the bad model subset has more distortion; In contrast, DBCNN is more balanced when analyzing data from different AGI models.
		
		Considering the difference in phrase length, we define the phrase as prompt 0-3 according to the number of `detail' and `style' according to the description of Sec. \ref {sec:analysis}. The larger the number, the more complex the phrase. Under the subset of different prompt lengths, the performance of the Perception model is shown in Tab. \ref{tab:perception} (b). Generally, the perceptual quality prediction performance of each model decreases to some extent as the prompts become longer. From the distribution in Fig. \ref{fig:Prompt}, it's believed that the decrease is a combined result of longer prompts making the content difficult to generate, and the perception model's insufficient predictive ability for low-quality content.
		
		For different styles in AGIs, due to the similarity of Abstract \& Sci-fi style (unpopular) and Anime \& Realistic Style (second-unpopular) shown in Fig. \ref{fig:Style}, we classify the styles as Tab. \ref{tab:perception} (c). The data show that the perceptual quality model works well in predicting the quality of AGI in unpopular styles, but not satisfactorily in popular styles.

		For T2I alignment, we conduct similar validation in Tab. \ref{tab:alignment} like perception. The result shows that the alignment model has a lot to improve when predicting the T2I alignment of images generated by the good AGI model, long prompts, and with popular styles. It is worth mentioning that due to the reasonable disassembly of the prompt, our StairReward far outperforms other methods in predicting the alignment of long prompts, thus taking the lead in the alignment index of the entire AGIQA-3K.
		
		\begin{table*}[tbph]
			\caption{Alignment metric performance results on the AGIQA-3K database and different subsets from different T2I AGI models. The best performance results are marked in {\bf\textcolor{red}{RED}} and the second performance results are marked in {\bf\textcolor{blue}{BLUE}}.}
			\label{tab:alignment}
			\centering
			\subtable[ All AGIQA-3K database and three different subsets from different T2I AGI model groups]{
				\begin{tabular}{c|ccc|ccc|ccc|ccc}
					\toprule
					\multicolumn{1}{c|}{\multirow{2}{*}{Metric}} & \multicolumn{3}{c|}{All} & \multicolumn{3}{c|}{Bad Model} & \multicolumn{3}{c|}{Medium Model} & \multicolumn{3}{c}{Good Model} \\ \cline{2-13} 
					\multicolumn{1}{c|}{}                        & SRoCC  & KRoCC  & PLCC   & SRoCC    & KRoCC    & PLCC     & SRoCC     & KRoCC     & PLCC      & SRoCC    & KRoCC    & PLCC     \\ \hline
					CLIP\cite{align:clip}                                         & 0.5972 & 0.4591 & 0.6839 & {\bf\textcolor{blue}{0.5463}}   & {\bf\textcolor{blue}{0.3833}}   & 0.5355   & 0.2272    & 0.1740    & 0.2916    & 0.2420   & 0.1837   & 0.3342   \\
					ImageReward\cite{database/align:ImageReward}                                  & {\bf\textcolor{blue}{0.7298}} & {\bf\textcolor{blue}{0.5390}} & {\bf\textcolor{blue}{0.7862}} & {\bf\textcolor{red}{0.5652}}   & {\bf\textcolor{red}{0.3965}}   & {\bf\textcolor{blue}{0.6869}}   & {\bf\textcolor{blue}{0.4464}}    & {\bf\textcolor{blue}{0.3058}}    & {\bf\textcolor{blue}{0.5109}}    & 0.3925   & 0.2686   & {\bf\textcolor{blue}{0.4966}}   \\
					HPS\cite{database/align:HPS}                                          & 0.6349 & 0.4580 & 0.7000 & 0.5255   & 0.3675   & 0.5803   & 0.2762    & 0.1865    & 0.3516    & 0.3126   & 0.2128   & 0.3498   \\
					PickScore\cite{database/align:PickAPic}                                    & 0.6977 & 0.5069 & 0.7633 & 0.4293   & 0.2944   & 0.5588   & 0.3962    & 0.2683    & 0.3924    & {\bf\textcolor{blue}{0.4183}}   & {\bf\textcolor{blue}{0.2898}}   & 0.4743   \\
					StairReward                                  & {\bf\textcolor{red}{0.7472}} & {\bf\textcolor{red}{0.5554}} & {\bf\textcolor{red}{0.8529}} & 0.5401   & 0.3775   & {\bf\textcolor{red}{0.7076}}   & {\bf\textcolor{red}{0.4642}}    & {\bf\textcolor{red}{0.3228}}    & {\bf\textcolor{red}{0.5423}}    & {\bf\textcolor{red}{0.4411}}   & {\bf\textcolor{red}{0.3086}}   & {\bf\textcolor{red}{0.5581}}  \\ \bottomrule
			\end{tabular}}
			
			\subtable[Four different AGIQA-3K subsets with different prompt length]{
				\begin{tabular}{c|ccc|ccc|ccc|ccc}
					\toprule
					\multicolumn{1}{c|}{\multirow{2}{*}{Metric}} & \multicolumn{3}{c|}{Prompt 0} & \multicolumn{3}{c|}{Prompt 1} & \multicolumn{3}{c|}{Prompt 3} & \multicolumn{3}{c}{Prompt 4} \\ \cline{2-13} 
					\multicolumn{1}{c|}{}                        & SRoCC    & KRoCC    & PLCC    & SRoCC    & KRoCC    & PLCC    & SRoCC    & KRoCC    & PLCC    & SRoCC    & KRoCC   & PLCC    \\ \hline
					CLIP\cite{align:clip}                                         & 0.6202   & 0.4738   & 0.7083  & 0.6174   & 0.4798   & 0.6867  & 0.6344   & 0.4937   & 0.6858  & 0.5618   & 0.4296  & 0.6447  \\
					ImageReward\cite{database/align:ImageReward}                                  & {\bf\textcolor{blue}{0.7678}}   & {\bf\textcolor{blue}{0.5717}}   & {\bf\textcolor{blue}{0.8031}}  & {\bf\textcolor{blue}{0.7216}}   & {\bf\textcolor{blue}{0.5341}}   & {\bf\textcolor{blue}{0.7873}}  & {\bf\textcolor{red}{0.7313}}   & {\bf\textcolor{blue}{0.5395}}   & {\bf\textcolor{blue}{0.7829}}  & {\bf\textcolor{blue}{0.6786}}   & {\bf\textcolor{blue}{0.4904}}  & {\bf\textcolor{blue}{0.7743}}  \\
					HPS\cite{database/align:HPS}                                          & 0.6623   & 0.4810   & 0.7008  & 0.6851   & 0.5006   & 0.7329  & 0.7032   & 0.5183   & 0.7541  & 0.5652   & 0.4025  & 0.7246  \\
					PickScore\cite{database/align:PickAPic}                                    & 0.7320   & 0.5389   & 0.7791  & 0.7084   & 0.5205   & 0.7778  & 0.6991   & 0.5074   & 0.7687  & 0.6264   & 0.4423  & 0.7176  \\
					StairReward                                  & {\bf\textcolor{red}{0.7682}}   & {\bf\textcolor{red}{0.5772}}   & {\bf\textcolor{red}{0.8468}}  & {\bf\textcolor{red}{0.7259}}   & {\bf\textcolor{red}{0.5387}}   & {\bf\textcolor{red}{0.8441}}  & {\bf\textcolor{blue}{0.7270}}   & {\bf\textcolor{red}{0.5396}}   & {\bf\textcolor{red}{0.8465}}  & {\bf\textcolor{red}{0.7312}}   & {\bf\textcolor{red}{0.5418}}  & {\bf\textcolor{red}{0.8713}}  \\ \bottomrule
			\end{tabular}}
			
			\subtable[Four different AGIQA-3K subsets with different style]{
				\begin{tabular}{c|ccc|ccc|ccc|ccc}
					\toprule
					\multicolumn{1}{c|}{\multirow{2}{*}{Metric}} & \multicolumn{3}{c|}{Abstract \& Sci-fi Style} & \multicolumn{3}{c|}{Anime \& Realistic Style} & \multicolumn{3}{c|}{Baroque Style} & \multicolumn{3}{c}{No Style} \\ \cline{2-13} 
					\multicolumn{1}{c|}{}                        & SRoCC    & KRoCC    & PLCC    & SRoCC    & KRoCC    & PLCC    & SRoCC    & KRoCC    & PLCC    & SRoCC    & KRoCC   & PLCC    \\ \hline
					CLIP\cite{align:clip}                                         & 0.6268   & 0.4982   & 0.6320  & 0.7064   & 0.5598   & 0.6997  & 0.5473   & 0.4171   & 0.7175  & 0.5814   & 0.4415  & 0.6706  \\
					ImageReward\cite{database/align:ImageReward}                                  & {\bf\textcolor{blue}{0.7476}}   & {\bf\textcolor{blue}{0.5557}}   & {\bf\textcolor{blue}{0.7692}}  & {\bf\textcolor{blue}{0.7558}}   & {\bf\textcolor{blue}{0.5634}}   & 0.8291  & 0.6689   & 0.4885   & {\bf\textcolor{blue}{0.7931}}  & {\bf\textcolor{blue}{0.7369}}   & {\bf\textcolor{blue}{0.5434}}  & {\bf\textcolor{blue}{0.7784}}  \\
					HPS\cite{database/align:HPS}                                          & 0.6203   & 0.4465   & 0.6512  & 0.6630   & 0.4790   & 0.7462  & 0.6640   & 0.4800   & 0.7529  & 0.6262   & 0.4502  & 0.6798  \\
					PickScore\cite{database/align:PickAPic}                                    & 0.6238   & 0.4500   & 0.6992  & {\bf\textcolor{red}{0.7651}}   & {\bf\textcolor{red}{0.5699}}   & {\bf\textcolor{blue}{0.8406}}  & {\bf\textcolor{red}{0.7114}}   & {\bf\textcolor{red}{0.5173}}   & 0.7614  & 0.7054   & 0.5139  & 0.7655  \\
					StairReward                                  & {\bf\textcolor{red}{0.7584}}   & {\bf\textcolor{red}{0.5698}}   & {\bf\textcolor{red}{0.8668}}  & 0.7396   & 0.5588   & {\bf\textcolor{red}{0.8811}}  & {\bf\textcolor{blue}{0.6760}}   & {\bf\textcolor{blue}{0.4972}}   & {\bf\textcolor{red}{0.8622}}  & {\bf\textcolor{red}{0.7447}}   & {\bf\textcolor{red}{0.5541}}  & {\bf\textcolor{red}{0.8363}}  \\ \bottomrule
			\end{tabular}}
		\end{table*}
		
		Generally, considering the performance of perception and alignment evaluation models, future work can be carried out in the following aspects:
		\begin{itemize}
			\item For both types of models, the existing perception and alignment models are good at distinguishing between excellent and poor quality AGI, but when faced with similar quality results from the same T2I AGI model, the assessment is not accurate enough. How to distinguish AGIs with similar subjective quality is the most urgent problem to be solved in future quality models.
			\item For perception models, the IQA models (especially the DL-based models) have excellent agreement with the HVS subjective score. Therefore, the future T2I AGI model can consider replacing the traditional loss function with a DL-based model and inspiring generation through perception.
			\item For alignment models, their performance has a certain gap with the above-mentioned perception assessment task. Our proposed StairReward improves the alignment assessment performance to some extent, but a more accurate quality model is still needed in the future.
		\end{itemize}

		\subsection{Ablation Study}
		
		To validate the contributions of the different modules in StairReward, we also conduct an ablation study and its results are shown in Tab. \ref{tab:abandon}. The factors are specified as (word): prompt segmentation and (image): image cutting. The results show that removing any single factor leads to performance degradation, which confirms that they all contribute to the performance results in Tab. \ref{tab:alignment}.
		
		\begin{table}[tbph]
			\caption{Abandoning different module in StairReward.}
			\label{tab:abandon}
			\centering
			\begin{tabular}{c|ccc}
				\toprule
				Ablation & SRoCC  & KRoCC  & PLCC   \\ \hline
				none                           & \textbf{0.7472} & \textbf{0.5554} & \textbf{0.8529} \\
				word                           & \textbf{0.7131} & \textbf{0.5203} & \textbf{0.6897} \\
				image                          & \textbf{0.7422} & \textbf{0.5532} & \textbf{0.7772} \\
				all                            & \textbf{0.7298} & \textbf{0.5390} & \textbf{0.7862} \\ \bottomrule
			\end{tabular}
		\end{table}
		
		
		\section{Conclusion}
		
		With the continuous advancement of deep learning technology, a large number of T2I AGI models have emerged in recent years. Different prompt inputs and parameter selections will lead to huge differences in AGI quality. Therefore, refinement and filtering are required before actual use. Therefore, there is an urgent need to develop objective models to assess the quality of AGI. In this paper, we first discuss important evaluation aspects, formulating subjective evaluation criteria in terms of perception/alignment. Then, we established the AGI quality assessment database AGIQA-3K, which covers the largest number of AGI models, the most fine-grained and the most layers, and contains 2,982 AGIs generated from the diffusion model. A well-organized subjective experiment is conducted to collect quality labels for AGI. Subsequently, benchmark experiments are conducted to evaluate the performance of current IQA models. Experimental results show that current perception/alignment models cannot handle the AGIQA task well, especially considering the limited performance of existing alignment models, we propose StairReward to objectively evaluate the alignment quality. In conclusion, both perception/alignment models need to be improved in the future, and AGIQA-3K lays the foundation for this improvement.

	\end{document}